\documentclass{article}

\usepackage{arxiv}

\usepackage[utf8]{inputenc}
\usepackage[T1]{fontenc}
\usepackage{hyperref}
\usepackage{url}
\usepackage{booktabs}
\usepackage{amsfonts}
\usepackage{nicefrac}
\usepackage{microtype}
\usepackage{graphicx}
\graphicspath{{./figures/}{./}}
\usepackage{enumitem}
\usepackage{multirow}
\usepackage{array}
\usepackage{amsmath}
\usepackage{xcolor}
\usepackage{colortbl}
\usepackage{tikz}
\usetikzlibrary{shapes.geometric,arrows.meta,positioning,fit,backgrounds}
\usepackage{float}
\usepackage{makecell}

\newcolumntype{C}[1]{>{\centering\arraybackslash}p{#1}}
\newcolumntype{L}[1]{>{\raggedright\arraybackslash}p{#1}}

\definecolor{goodgreen}{RGB}{26,122,26}
\definecolor{badred}{RGB}{192,0,0}
\definecolor{pipeblue}{RGB}{173,216,230}
\definecolor{pipeorange}{RGB}{255,200,120}
\definecolor{pipegreen}{RGB}{144,238,144}
\definecolor{pipegray}{RGB}{220,220,220}

\tikzset{
  pipebox/.style={rectangle,rounded corners=4pt,minimum width=2.2cm,
                  minimum height=0.85cm,text centered,draw=black,
                  font=\small,line width=0.8pt,align=center},
  arrow/.style={-Stealth,thick},
  groupbox/.style={rectangle,rounded corners=6pt,draw=gray!60,
                   dashed,line width=0.8pt,inner sep=8pt}
}

\title{A Benchmark of Classical and Deep Learning Models for Agricultural Commodity Price Forecasting on A Novel Bangladeshi Market Price Dataset}

\author{
  Tashreef Muhammad \\
  Department of Computer Science and Engineering\\
  Southeast University, Dhaka, Bangladesh\\
  \texttt{tashreef.muhammad@seu.edu.bd}
  \And
  Tahsin Ahmed \\
  Department of Computer Science and Engineering\\
  University of Dhaka, Bangladesh
  \And
  Meherun Farzana \\
  Department of Computer Science and Engineering\\
  University of Dhaka, Bangladesh
  \And
  Md.\ Mahmudul Hasan \\
  Department of Computer Science and Engineering\\
  University of Dhaka, Bangladesh
  \And
  Abrar Eyasir \\
  Department of Computer Science and Engineering\\
  University of Dhaka, Bangladesh
  \And
  Md.\ Emon Khan \\
  Department of Computer Science and Engineering\\
  University of Dhaka, Bangladesh
  \And
  Mahafuzul Islam Shawon \\
  Department of Computer Science and Engineering\\
  University of Dhaka, Bangladesh
  \And
  Ferdous Mondol \\
  Department of Computer Science and Engineering\\
  University of Dhaka, Bangladesh
  \And
  Mahmudul Hasan \\
  Department of Computer Science and Engineering\\
  University of Dhaka, Bangladesh
  \And
  Muhammad Ibrahim\thanks{Corresponding author.} \\
  Department of Computer Science and Engineering\\
  University of Dhaka, Bangladesh\\
  \texttt{ibrahim313@du.ac.bd}
}

\begin{document}
\maketitle

\begin{abstract}
Accurate short-term forecasting of agricultural commodity prices is critical for food security
planning, market policy, and smallholder income stabilisation in developing economies, yet
publicly available machine-learning-ready datasets for this purpose remain scarce in South Asia.
This paper makes two primary contributions. First, we introduce AgriPriceBD, a novel benchmark dataset of
1,779 daily retail mid-prices for five key Bangladeshi commodities---garlic, chickpea, green
chilli, cucumber, and sweet pumpkin---spanning July~2020 to June~2025, extracted from government
market monitoring reports via an LLM-assisted digitisation pipeline and released publicly to
support reproducible research. Second, using this dataset we conduct a systematic comparative
evaluation of seven forecasting approaches spanning classical models---na\"{i}ve persistence,
SARIMA, and Prophet---and deep learning architectures---BiLSTM, a vanilla Transformer, a
Time2Vec-enhanced Transformer, and Informer---reporting both point accuracy
and Diebold-Mariano statistical significance tests, except for Informer as it produced erratic,
poorly-calibrated predictions on all commodities. We find that commodity price forecastability is
fundamentally heterogeneous: na\"{i}ve persistence dominates on near-random-walk commodities.
Contrary to expectations, learnable Time2Vec temporal encoding provides no statistically significant
advantage over fixed sinusoidal encoding on any commodity at this training scale, and causes
catastrophic degradation on the most volatile commodity (green chilli, $+146.1\%$ MAE,
$p<0.001$)---a practically important negative result for agricultural ML practitioners. Prophet
fails systematically across all commodities, a finding we attribute to the discrete step-function
price dynamics characteristic of developing-economy retail markets. The Informer architecture
produces erratic, poorly-calibrated predictions (prediction variance up to $50{\times}$
ground-truth on some commodities), confirming that sparse-attention Transformers require
substantially larger training sets than small-sample agricultural monitoring contexts can provide.
All code, models, and data are released for public use to enable direct replication and extension.
This research is expected to support policymakers, smallholder farmers, and food security agencies
in making informed, forward-looking market intervention decisions, and to serve as a reproducible
baseline for future forecasting research on agricultural commodity markets in Bangladesh and
similar developing economies.
\end{abstract}

\keywords{Agricultural price forecasting \and Benchmark dataset \and Transformer \and
Time2Vec \and Bangladesh \and Deep learning \and Food security \and Time series}

\section{Introduction}
\label{sec:intro}

Food price volatility poses a persistent challenge to food security, household welfare, and
macroeconomic stability across South Asia. In Bangladesh, a nation of over 177~million people
where food constitutes a large share of household expenditure
\cite{herteux2024forecasting, fao2023food}, anticipating near-term retail price movements for
key agricultural commodities has direct practical consequences. Farmers benefit from
forward-looking price signals when planning planting and sales decisions; policymakers require
reliable forecasts to activate market intervention mechanisms before supply shocks cascade to
consumers; traders and distributors can reduce post-harvest waste through improved logistics.
\cite{patil2023forecasting} explicitly frames accurate agricultural price forecasting as an
enabler of Sustainable Development Goal~2 (Zero Hunger), motivating the development of
forecasting infrastructure in food-insecure geographies.

Yet despite these stakes, the quantitative forecasting of Bangladeshi agricultural retail prices
remains largely unstudied in the machine learning literature. \textbf{Two gaps} are particularly
acute. \textbf{First}, no publicly available daily multi-commodity retail price benchmark exists
for Bangladesh; research has been confined to single commodities (typically rice) and classical
statistical methods \cite{hassan2013forecasting, hasan2020ascertaining, imran2022harnessing}.
\textbf{Second}, because Bangladeshi commodity prices exhibit discrete step-function
dynamics---extended periods of stability punctuated by sudden jumps---it is unclear whether
approaches designed for smooth time series transfer to this setting. In particular, the
widely-used Prophet framework \cite{taylor2018forecasting} and large-scale Transformer
architectures such as Informer \cite{zhou2021informer} have not been evaluated under these
conditions.

This paper addresses both gaps. Our contributions are:
\begin{enumerate}[label=\roman*),leftmargin=*]
  \item \textbf{A novel benchmark dataset (AgriPriceBD).} We release daily retail mid-prices
    for five Bangladeshi agricultural commodities spanning five years, extracted from government PDF
    reports via an LLM-assisted pipeline. To the best of our knowledge this is the first
    publicly available daily multi-commodity retail price dataset for Bangladesh.
  \item \textbf{A systematic comparative evaluation.} We evaluate seven forecasting approaches
    on this dataset, including two architectures---Prophet and Informer---that have not
    previously been tested on discrete step-function retail price series in developing-economy
    settings. We document their failure modes explicitly.
  \item \textbf{A controlled temporal encoding ablation with statistical significance testing.}
    We isolate the contribution of learnable Time2Vec temporal embeddings against fixed
    sinusoidal positional encoding using the Diebold-Mariano test, providing evidence for when
    learnable temporal representations are and are not beneficial.
\end{enumerate}

Our central finding is that commodity price forecastability is fundamentally heterogeneous. No
single model dominates across all commodities, and the signal-to-noise structure of a
commodity's price series---rather than model complexity---is the primary determinant of
forecasting accuracy. The released AgriPriceBD dataset (Mendeley Data: \url{https://data.mendeley.com/datasets/bkmxnrn3hn}) and codebase
(\url{https://github.com/TashreefMuhammad/Bangladesh-Agri-Price-Forecast}) are intended as
infrastructure for future work, enabling researchers to extend, replicate, and build on
these baseline results.

Section~\ref{sec:related} reviews related work. Section~\ref{sec:method} describes the dataset
and experimental design. Section~\ref{sec:results} presents results. Section~\ref{sec:discussion}
discusses findings. Section~\ref{sec:conclusion} concludes.

\section{Related Work}
\label{sec:related}

\subsection{Classical and Statistical Forecasting}

Autoregressive time series models have served as the standard baseline for agricultural price
forecasting for decades. \cite{box2015time} established the ARIMA framework, and
SARIMA---its seasonal extension---remains competitive on commodity series with stable periodic
structure \cite{hyndman2018forecasting}. In developing economies, SARIMA has been applied to
rice prices in Bangladesh \cite{hassan2013forecasting}, vegetable prices in India
\cite{dasari2025price}, and pulse markets across South Asia \cite{mishra2021state}. Its
principal limitation is linearity, which fails when prices exhibit nonlinear structural breaks
or discrete jumps.

Prophet \cite{taylor2018forecasting} decomposes time series into smooth trend, seasonality, and
holiday components using a piecewise-linear model, and has been widely adopted in applied
forecasting due to its interpretability. However, its smoothness assumptions are violated by
the discrete step-function price dynamics that characterise developing-economy retail
markets---a point this paper demonstrates empirically and discusses in Section~\ref{sec:prophet}.

\subsection{Deep Learning for Agricultural Price Forecasting}

The study of deep learning models for agricultural price forecasting has been quite prevailing
in recent years \cite{paul2025deep, aslam2024nbeats}. Long short-term memory networks
\cite{hochreiter1997long} provided the first effective deep learning approach to sequential
modelling. Bidirectional variants have shown improved performance on agricultural and commodity
price series across multiple geographies \cite{bahar2024dual, nensi2025implementing,
dasari2025price}. \cite{manogna2025enhancing, singh2025cherry} conducted a comprehensive deep
learning comparison for agricultural commodity price forecasting in India, finding that ensemble
and recurrent approaches outperform classical baselines on volatile series.
\cite{sari2024various} compared optimised machine learning techniques across multiple commodity
markets, documenting substantial variation in model performance across commodities---consistent
with the heterogeneous forecastability finding reported here. \cite{patil2023forecasting}
emphasised the practical connection between forecasting accuracy and food security outcomes in
developing economies.

The Transformer architecture \cite{vaswani2017attention} replaced recurrence with multi-head
self-attention. Informer \cite{zhou2021informer} extended it to very long horizons via sparse
attention, designed for industrial datasets with 10{,}000+ observations. PatchTST
\cite{nie2022time} introduced patch-based tokenisation for time series Transformers,
substantially improving efficiency on large benchmarks. As we demonstrate, these large-scale
architectures require substantially more training data than small-sample agricultural monitoring
contexts typically provide.

\subsection{Temporal Encoding and Learnable Representations}

Fixed sinusoidal positional encoding \cite{vaswani2017attention} communicates relative sequence
order but carries no information about the absolute temporal position of an observation within
a seasonal cycle---a critical limitation for harvest-cycle-driven agricultural prices.
\cite{kazemi2019time2vec} proposed Time2Vec, a learnable temporal embedding that combines a
linear trend term with learned sinusoidal functions at discovered frequencies. This allows the
model to identify dominant periodicities from data rather than assuming them a priori.
\cite{muhammad2023transformer} previously applied a Transformer-based model to the Bangladesh
stock market, establishing the feasibility of attention architectures in the Bangladeshi
context; the present work extends this to agricultural retail forecasting with an explicit
ablation of temporal encoding.

\subsection{Agricultural Forecasting in Bangladesh and South Asia}

Though there are available datasets in other countries (e.g.\ India) \cite{agmarknet},
existing work on Bangladeshi commodity markets is narrow in scope.
\cite{hassan2013forecasting} applied SARIMA to wholesale rice prices.
\cite{hasan2020ascertaining} used machine learning for rice price fluctuation analysis, limited
to a single commodity. \cite{imran2022harnessing} incorporated meteorological covariates into
rice price prediction, demonstrating the potential value of exogenous features.
\cite{islam2024comparative} applied ML to Aman rice yields, addressing the production side
rather than retail prices. The absence of a publicly available daily multi-commodity retail
benchmark has likely constrained research activity in this geography.

Across South Asia more broadly, \cite{dasari2025price} applied SARIMA-LSTM to vegetable price
forecasting in India, and \cite{mishra2021state} analysed pulse production trends using ARIMA.
In other developing-economy settings, \cite{nensi2025implementing} applied LSTM to
high-volatility food commodity prices in Indonesia, and \cite{sari2024various} optimised ML
approaches for commodity price prediction in Turkey. None of these studies addresses the
Bangladeshi retail market or provides a reusable daily multi-commodity benchmark.

\subsection{Gap Analysis}
\label{sec:gap}

Table~\ref{tab:gap} synthesises the characteristics of closely related studies and identifies
the specific gaps addressed by this work.

\begin{table}[htbp]
\centering
\caption{Gap analysis of related work. BD~=~Bangladesh; IN~=~India; MY~=~Malaysia;
ID~=~Indonesia; TR~=~Turkey. \checkmark~=~present; --~=~absent.}
\label{tab:gap}
\setlength{\tabcolsep}{4pt}
\renewcommand{\arraystretch}{1.15}
\begin{tabular}{L{3.5cm} C{0.7cm} L{1.8cm} C{1.5cm} C{1.5cm} C{1.5cm} C{1.5cm}}
\toprule
\textbf{Study} & \textbf{Geo.} & \textbf{Commodity} &
\textbf{DL} & \textbf{Learn.\ Temp.} &
\textbf{Multi-commod.} & \textbf{BD Public Retail} \\
\midrule
Hassan et al.\ \cite{hassan2013forecasting}    & BD & Rice       & --         & -- & -- & -- \\
Hasan et al.\ \cite{hasan2020ascertaining}     & BD & Rice       & \checkmark & -- & -- & -- \\
Imran et al.\ \cite{imran2022harnessing}       & BD & Rice       & \checkmark & -- & -- & -- \\
Islam et al.\ \cite{islam2024comparative}      & BD & Rice yield & \checkmark & -- & -- & -- \\
Bahar et al.\ \cite{bahar2024dual}             & MY & Palm oil   & \checkmark & -- & -- & -- \\
Nensi et al.\ \cite{nensi2025implementing}     & ID & Vegetables & \checkmark & -- & \checkmark & -- \\
Dasari et al.\ \cite{dasari2025price}           & IN & Vegetables & \checkmark & -- & \checkmark & -- \\
Manogna et al.\ \cite{manogna2025enhancing}    & IN & Agri.      & \checkmark & -- & \checkmark & -- \\
Sari et al.\ \cite{sari2024various}            & TR & Agri.      & \checkmark & -- & \checkmark & -- \\
Muhammad et al.\ \cite{muhammad2023transformer}& BD & Stock      & \checkmark & -- & -- & -- \\
\midrule
\textbf{This work} & \textbf{BD} & \textbf{5 agri.} &
\checkmark & \checkmark & \checkmark & \checkmark \\
\bottomrule
\end{tabular}
\vspace{2pt}
\begin{flushleft}
\small\textit{\textbf{BD Public Retail}: first publicly available daily multi-commodity retail
price dataset for Bangladesh. \textbf{Learn.\ Temp.}: learnable temporal encoding (e.g.,
Time2Vec) applied to agricultural retail price forecasting in this geography.}
\end{flushleft}
\end{table}

\section{Data and Methodology}
\label{sec:method}

\subsection{AgriPriceBD: Dataset Construction and LLM-Assisted Extraction Pipeline}
\label{sec:dataset}

\paragraph{Data source.}
The Bangladesh government market monitoring system publishes daily PDF reports recording
minimum and maximum retail prices in Bangladeshi Taka (BDT) per kilogram for agricultural
commodities at monitored markets. Five commodities were selected based on nutritional
significance and consumption prevalence: garlic, chickpea, green chilli, cucumber, and sweet
pumpkin. The extraction covers 22~July~2020 to 4~June~2025, yielding 1,779 daily observations
per commodity. AgriPriceBD is deposited on Mendeley Data (\url{https://data.mendeley.com/datasets/bkmxnrn3hn})
and all code is available at
\url{https://github.com/TashreefMuhammad/Bangladesh-Agri-Price-Forecast}.

\begin{figure}[htbp]
\centering
\begin{tikzpicture}[node distance=0.65cm and 1.6cm]
  \node[pipebox, fill=pipeblue!60]   (portal)  {Govt.\ Portal\\ (PDF reports)};
  \node[pipebox, fill=pipeblue!40,   below=of portal]  (dl)     {Systematic\\ Download};
  \node[pipebox, fill=pipeorange!70, right=1.6cm of portal] (llm)    {Gemini API\\ (LLM parsing)};
  \node[pipebox, fill=pipeorange!40, below=of llm]     (prompt) {Bilingual Prompt\\ (EN + Bangla)};
  \node[pipebox, fill=pipegray!80,   right=1.6cm of llm]   (val)    {Range \& Date\\ Validation};
  \node[pipebox, fill=pipegray!60,   below=of val]     (clean)  {Merge \&\\ Clean CSV};
  \node[pipebox, fill=pipegreen!70,  right=1.6cm of val]   (mid)    {Mid-price\\ Computation};
  \node[pipebox, fill=pipegreen!50,  below=of mid]     (out)    {\textbf{Final Dataset}\\ 1{,}779 obs/commod.};
  \draw[arrow] (portal) -- (dl);
  \draw[arrow] (dl)     -- node[above, font=\tiny]{PDF}    (llm);
  \draw[arrow] (llm)    -- (prompt);
  \draw[arrow] (prompt) -- node[above, font=\tiny, pos=0.25]{JSON} (val);
  \draw[arrow] (val)    -- (clean);
  \draw[arrow] (clean)  -- node[above, font=\tiny]{CSV}    (mid);
  \draw[arrow] (mid)    -- (out);
  \node[font=\footnotesize\bfseries, above=0.12cm of portal] (s1) {Stage 1};
  \node[font=\footnotesize\bfseries, above=0.12cm of llm]    (s2) {Stage 2};
  \node[font=\footnotesize\bfseries, above=0.12cm of val]    (s3) {Stage 3};
  \node[font=\footnotesize\bfseries, above=0.12cm of mid]    (s4) {Stage 4};
  \begin{scope}[on background layer]
    \node[groupbox, fit=(s1)(portal)(dl)]  {};
    \node[groupbox, fit=(s2)(llm)(prompt)] {};
    \node[groupbox, fit=(s3)(val)(clean)]  {};
    \node[groupbox, fit=(s4)(mid)(out)]    {};
  \end{scope}
\end{tikzpicture}
\caption{LLM-assisted dataset extraction pipeline. Daily government PDF market reports are
parsed via the Gemini API using bilingual structured prompts (English and Bangla commodity name
synonyms), validated against domain constraints, and aggregated into per-commodity CSV files.
The forecast target is the retail mid-price: $p_t = (\mathrm{min}_t + \mathrm{max}_t)/2$.}
\label{fig:pipeline}
\end{figure}
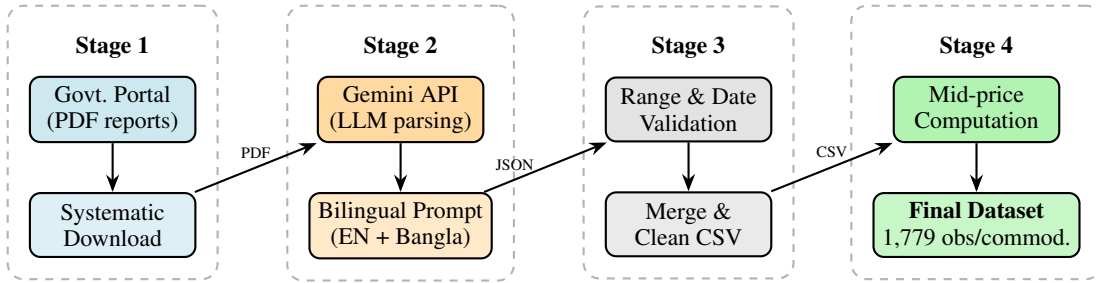

\paragraph{Extraction pipeline.}
Because no structured digital API exists, an LLM-assisted extraction pipeline was developed.
Figure~\ref{fig:pipeline} illustrates the four-stage process.

\begin{enumerate}[label=\textbf{Stage~\arabic*:},leftmargin=*]
  \item \textbf{PDF retrieval.} Daily reports were systematically downloaded from the
    government portal. Reports are non-standardised across years, with varying table layouts,
    column ordering, and text encoding.
  \item \textbf{LLM-assisted parsing.} Each PDF was passed to the Gemini API with a structured
    prompt requesting minimum and maximum retail prices per commodity in JSON format. Bilingual
    commodity name synonyms in English and Bangla handled transliteration variation across
    report years.
  \item \textbf{Validation and cleaning.} Extracted records were validated against price-range
    constraints (flagging values outside 0.1--500~BDT/kg), checked for date continuity, and
    merged into unified per-commodity CSV files.
  \item \textbf{Mid-price computation.} The forecast target was computed as
    $p_t = (\text{min}_t + \text{max}_t)/2$, providing a signal less sensitive to daily
    boundary fluctuations than either extreme alone.
\end{enumerate}

\paragraph{Data quality.}
Four records in the green chilli series (22, 23, 24, and 29~January~2024) contain zero-valued
prices inconsistent with surrounding observations ($\approx$62--70~BDT/kg), attributed to
government portal outages. These represent 0.22\% of the series and were retained as-is with
documentation for transparency.

\begin{figure}[htbp]
  \centering
  \includegraphics[width=0.55\textwidth]{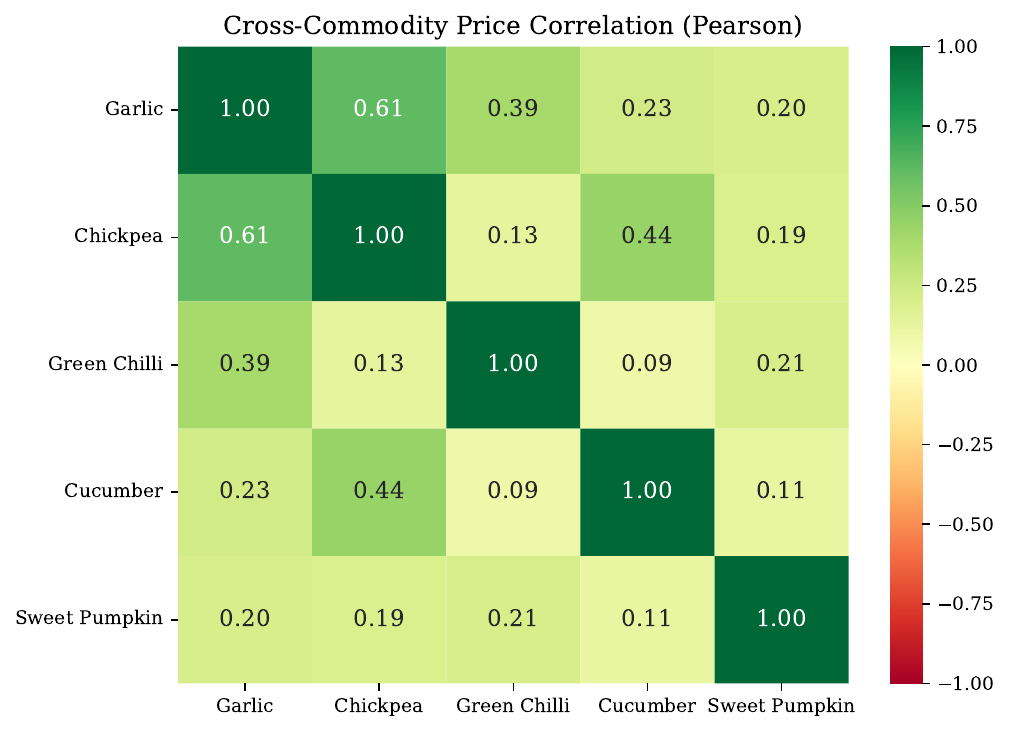}
  \caption{Cross-commodity Pearson correlation matrix of daily retail mid-prices (July~2020--June~2025). Garlic--chickpea show the strongest positive correlation ($r=0.61$), reflecting shared import-parity pricing dynamics. Cucumber shows low correlation with green chilli ($r=0.09$) and sweet pumpkin ($r=0.11$), but a moderate correlation with chickpea ($r=0.44$).}
  \label{fig:corr}
\end{figure}

\paragraph{Cross-commodity structure.}
Figure~\ref{fig:corr} presents the Pearson correlation matrix across all five commodity mid-price series. Garlic and chickpea exhibit the strongest co-movement ($r=0.61$), consistent with both being imported staples subject to common import policy dynamics. Cucumber shows the weakest correlations with green chilli ($r=0.09$) and sweet pumpkin ($r=0.11$), though a moderate correlation with chickpea ($r=0.44$) suggests some shared supply dynamics. Overall cross-commodity correlations are sufficiently low to support univariate modelling as a meaningful baseline.

\paragraph{Summary statistics and stationarity.}
Table~\ref{tab:dataset} reports summary statistics and Augmented Dickey-Fuller (ADF)
stationarity test results. Garlic and chickpea are non-stationary, reflecting multi-year price
trends. Green chilli, cucumber, and sweet pumpkin are stationary. This heterogeneity motivates
the inclusion of both differencing-based (SARIMA) and level-based (deep learning) models.

\begin{table}[htbp]
\centering
\caption{Dataset summary statistics (July~2020 -- June~2025, prices in BDT/kg).}
\label{tab:dataset}
\setlength{\tabcolsep}{7pt}
\renewcommand{\arraystretch}{1.1}
\begin{tabular}{L{2.4cm} C{1.1cm} C{0.9cm} C{0.9cm}
                C{0.9cm} C{0.9cm} C{1.4cm} C{1.2cm}}
\toprule
\textbf{Commodity} & \textbf{ADF \textit{p}} &
\textbf{Min} & \textbf{Max} & \textbf{Mean} & \textbf{Std} &
\textbf{Stationary?} & \textbf{R/S} \\
\midrule
Garlic        & 0.428      & 39.5 & 325.0 & 119.1 & 69.1 & No  & 0.93 \\
Chickpea      & 0.607      & 72.5 & 142.5 & 90.6 & 17.8 & No  & 1.32 \\
Green Chilli  & 0.003      & 0.0  & 260.0 & 87.3 & 56.2 & Yes & 0.74 \\
Cucumber      & ${<}0.001$ & 19.0 & 115.0 & 45.8 & 15.8 & Yes & 1.23 \\
Sweet Pumpkin & 0.008      & 11.5 &  52.5 & 25.3 &  7.3 & Yes & 0.70 \\
\bottomrule
\end{tabular}
\vspace{2pt}
\begin{flushleft}
\small\textit{$N=1{,}779$ daily observations per commodity.
ADF: Augmented Dickey-Fuller $p$-value; stationary if $p<0.05$.
R/S: residual-to-seasonal standard deviation ratio from STL decomposition.
Higher R/S indicates greater dominance of unpredictable noise over exploitable periodicity.}
\end{flushleft}
\end{table}

\subsection{Preprocessing and Evaluation Protocol}

Daily prices were forward-filled for any isolated missing dates. Each commodity was processed
independently as a univariate time series, a design choice supported by the low cross-commodity
correlation observed in Figure~\ref{fig:corr}. A strict temporal split was applied: 80\%
training (1,423~days), 10\% validation (178~days), 10\% test (178~days). No shuffling was
applied; temporal ordering was preserved throughout to prevent information leakage from future
observations. This protocol is consistent with established time series evaluation practice
\cite{hyndman2018forecasting}.

Standard $k$-fold cross-validation is inappropriate for time series data due to temporal
dependencies and look-ahead bias \cite{hyndman2018forecasting}. The test period (May--June~2025)
represents a genuine out-of-sample evaluation on the most recently available data.
Walk-forward validation over multiple windows is recommended for future work on extended
datasets.

Normalisation used MinMax \cite{goodfellow2016deep} scaling fit exclusively on the training
split. All reported metrics are computed on inverse-transformed (original-scale) predictions.
Sliding windows of length 90~days were used to construct model inputs, each producing a 14-day
forecast.

\subsection{Models}
\label{sec:models}

Seven forecasting approaches were evaluated, spanning two broad families: \textbf{classical
models}---Na\"{i}ve Persistence, SARIMA, and Prophet---which rely on statistical or
decomposition-based formulations without neural network components; and \textbf{deep learning
architectures}---BiLSTM, Vanilla Transformer, T2V-Transformer, and Informer---which learn
representations directly from data. All deep learning models used Adam optimisation
\cite{kingma2014adam}, initial learning rate $1\times10^{-3}$, Huber loss,
ReduceLROnPlateau scheduling (factor 0.5, patience 10), early stopping (patience 20), and
maximum 150~epochs. Random seed 42 was used throughout.

\paragraph{Na\"{i}ve Persistence}
Predicts the next 14~days as equal to the last observed price. Zero parameters; serves as the
floor baseline for assessing whether model complexity is justified.

\paragraph{SARIMA}
Seasonal ARIMA fit per commodity using the Hyndman-Khandakar automatic order selection
algorithm via pmdarima \cite{hyndman2018forecasting}, with weekly seasonal period $m=7$.
Rolling expanding-window evaluation over the test period.

\paragraph{Prophet}
Configured with Bangladesh-specific holidays (Ramadan, Eid ul-Fitr, Eid ul-Adha, 2020--2025).
Default seasonality settings. Prophet's failure mode is analysed in
Section~\ref{sec:prophet}.

\paragraph{BiLSTM}
Two-layer bidirectional LSTM \cite{hochreiter1997long}, hidden dimension 64, dropout 0.1
(garlic, chickpea, cucumber) and 0.3 (green chilli, sweet pumpkin).

\paragraph{Informer (preliminary, excluded from main comparison)}
\label{sec:informer_method}
The Informer \cite{zhou2021informer} was evaluated using its standard configuration
(e-layers=2, d-layers=1, $d_\text{model}$=64, 4~heads, $d_\text{ff}$=256, factor=5).
Results are reported separately in Section~\ref{sec:informer_results}.

\paragraph{Vanilla Transformer}
Two Pre-LayerNorm encoder layers; 4 attention heads; $d_\text{model}=64$; $d_\text{ff}=256$;
dropout~$=0.1$ (garlic, chickpea, cucumber) and 0.3 (green chilli, sweet pumpkin). Input
sequences are projected from the univariate price signal to $d_\text{model}$ via a linear
layer, then summed with fixed sinusoidal positional encodings \cite{vaswani2017attention}.
The last token passes through a linear head to produce the 14-day forecast. Pre-LayerNorm
was adopted for training stability on small datasets.

\begin{figure}[htbp]
  \centering
  \includegraphics[width=0.9\textwidth]{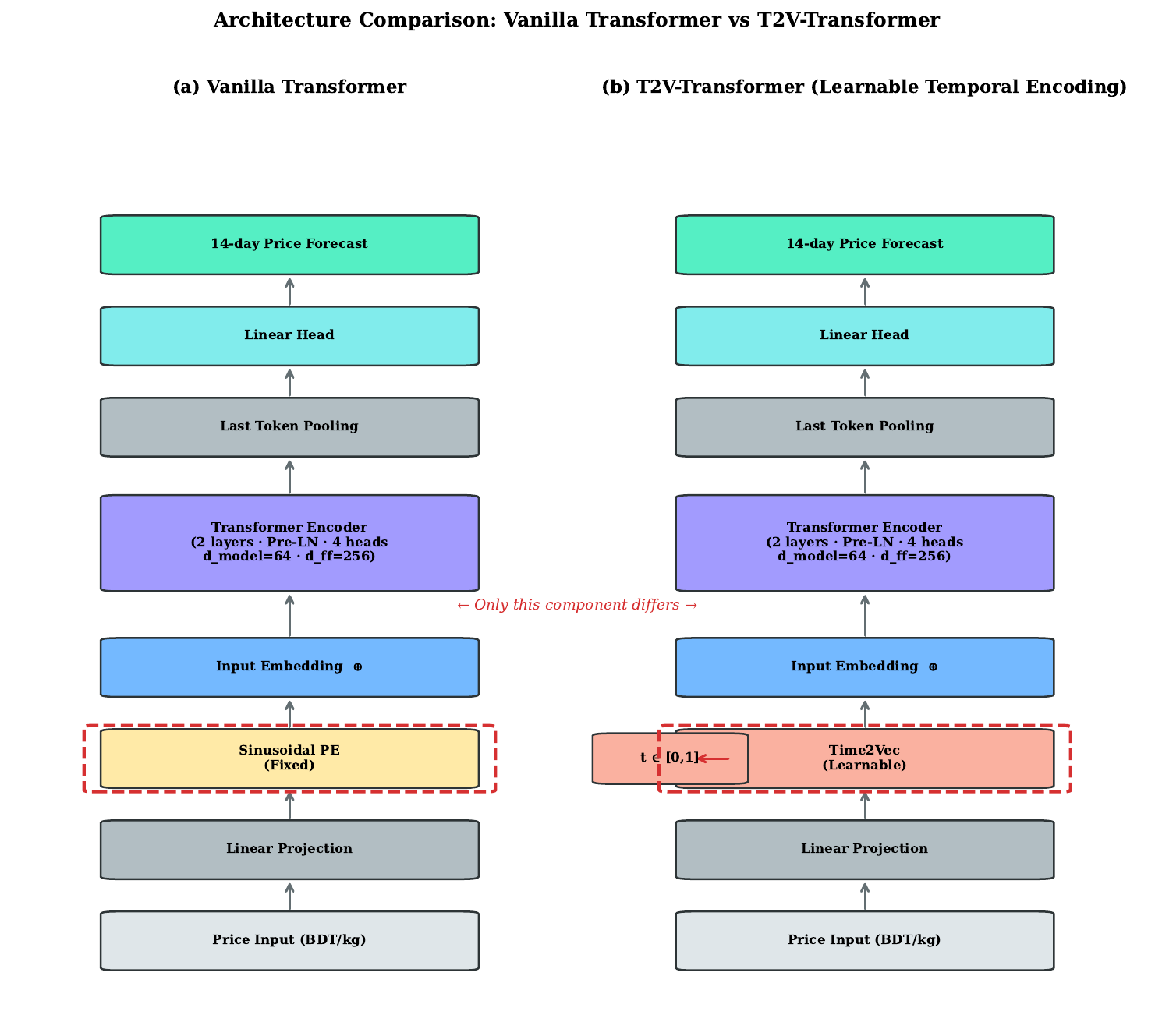}
  \caption{Architecture comparison. \textbf{(a)}~Vanilla Transformer uses fixed sinusoidal
  positional encoding computed from sequence position. \textbf{(b)}~T2V-Transformer replaces
  this with Time2Vec learnable temporal embeddings parameterised by global normalised time
  $\tau\in[0,1]$. All other components are held constant across both variants, enabling a
  clean controlled ablation of the temporal encoding contribution.}
  \label{fig:arch}
\end{figure}

\paragraph{T2V-Transformer (Time2Vec-Enhanced Transformer)}
Architecturally identical to the vanilla Transformer with one modification: fixed sinusoidal
positional encodings are replaced by Time2Vec learnable temporal embeddings
\cite{kazemi2019time2vec}. Figure~\ref{fig:arch} illustrates the architecture comparison.
We emphasise that Time2Vec is an existing method proposed by \cite{kazemi2019time2vec};
the T2V-Transformer here serves as an ablation target to determine whether learnable temporal
encoding improves upon fixed sinusoidal PE in this small-sample agricultural setting, rather
than as a methodological contribution in its own right.

Time2Vec maps a scalar time input $\tau$ to a $k$-dimensional embedding:
\begin{equation}
\text{T2V}(\tau)[i] = \begin{cases}
  \omega_i\tau + \varphi_i & \text{if } i = 0 \\[4pt]
  \sin(\omega_i\tau + \varphi_i) & \text{if } 1 \le i \le k{-}1
\end{cases}
\label{eq:t2v}
\end{equation}
where $\omega_i$ and $\varphi_i$ are learnable parameters. We use $k=32$, with frequencies
initialised on a logarithmic scale ($0.01$--$10.0$) to encourage discovery of both weekly and
seasonal cycles. The time index $\tau$ is the global position of each observation in the full
five-year series normalised to $[0,1]$, enabling the model to learn inter-year seasonal
patterns rather than within-window relative positions. The 32-dimensional output is projected
to $d_\text{model}=64$ via a linear layer before summation with value embeddings.

Commodity-specific dropout regularisation was applied based on validation loss dynamics:
dropout~$=0.3$ for green chilli and sweet pumpkin (both exhibiting rising validation loss
under the default 0.1), and dropout~$=0.1$ for other commodities. This override was applied
consistently to all three deep learning models.

\begin{table}[htbp]
\centering
\caption{Hyperparameter settings for all models. DL models share the training protocol in
the lower block. ``Auto'' = automatically selected by pmdarima's Hyndman-Khandakar algorithm
per commodity.}
\label{tab:hyperparams}
\setlength{\tabcolsep}{5pt}
\renewcommand{\arraystretch}{1.15}
\begin{tabular}{L{3.5cm} L{4.5cm} L{7.0cm}}
\toprule
\textbf{Model} & \textbf{Hyperparameter} & \textbf{Value} \\
\midrule
\multirow{3}{*}{SARIMA}
  & Order $(p,d,q)$            & Auto (AIC minimisation, pmdarima) \\
  & Seasonal order $(P,D,Q,m)$ & Auto, $m=7$ (weekly period) \\
  & Evaluation                 & Rolling expanding window \\
\midrule
\multirow{3}{*}{Prophet}
  & Yearly seasonality         & Enabled \\
  & Weekly seasonality         & Enabled \\
  & Holidays                   & BD-specific (Eid ul-Fitr, Eid ul-Adha,
                                  Ramadan; 2020--2025) \\
\midrule
\multirow{4}{*}{BiLSTM}
  & Layers / hidden dim        & 2 / 64 \\
  & Dropout                    & 0.1 (garlic, chickpea, cucumber);
                                  0.3 (green chilli, sweet pumpkin) \\
  & Sequence length / horizon  & 90 / 14 \\
  & Parameters (approx.)       & $\approx$134{,}000 \\
\midrule
\multirow{7}{*}{Vanilla Transformer}
  & $d_\text{model}$ / heads       & 64 / 4 \\
  & Encoder layers / $d_\text{ff}$ & 2 / 256 \\
  & Positional encoding            & Fixed sinusoidal \\
  & Dropout                        & 0.1 (garlic, chickpea, cucumber);
                                      0.3 (green chilli, sweet pumpkin) \\
  & Norm                           & Pre-LayerNorm \\
  & Sequence length / horizon      & 90 / 14 \\
  & Parameters (approx.)           & $\approx$136{,}000 \\
\midrule
\multirow{8}{*}{T2V-Transformer}
  & $d_\text{model}$ / heads       & 64 / 4 \\
  & Encoder layers / $d_\text{ff}$ & 2 / 256 \\
  & Temporal encoding              & Time2Vec learnable ($k=32$) \\
  & T2V freq.\ initialisation      & Log-scale ($0.01$--$10.0$) \\
  & Time index $\tau$              & Global normalised $[0,1]$ \\
  & Dropout                        & 0.1 (garlic, chickpea, cucumber);
                                      0.3 (green chilli, sweet pumpkin) \\
  & Sequence length / horizon      & 90 / 14 \\
  & Parameters (approx.)           & $\approx$139{,}000 \\
\midrule
\multirow{7}{*}{\makecell[l]{\textit{Shared DL training} \\ \textit{protocol (BiLSTM,} \\
\textit{Vanilla Transformer,} \\ \textit{T2V-Transformer)}}}
  & Optimiser        & Adam \cite{kingma2014adam}, $\text{lr}=10^{-3}$ \\
  & Loss function    & Huber loss \\
  & LR schedule      & ReduceLROnPlateau (factor 0.5, patience 10) \\
  & Early stopping   & Patience 20 (val loss) \\
  & Max epochs / batch size & 150 / 32 \\
  & Normalisation    & MinMax scaling (fit on train only) \\
  & Random seed      & 42 \\
\bottomrule
\end{tabular}
\end{table}

\subsection{Evaluation Metrics}

\begin{align}
\text{MAE}  &= \frac{1}{n}\sum_{i=1}^n |y_i - \hat{y}_i| \\[4pt]
\text{RMSE} &= \sqrt{\frac{1}{n}\sum_{i=1}^n (y_i-\hat{y}_i)^2} \\[4pt]
\text{MAPE} &= \frac{100}{n}\sum_{i=1}^n
  \left|\frac{y_i-\hat{y}_i}{y_i}\right|
\end{align}

MAE provides an interpretable absolute error in BDT/kg. RMSE penalises large errors more
heavily and is sensitive to price spike episodes. MAPE enables cross-commodity comparison on
a relative scale.

\subsection{Statistical Significance: Diebold-Mariano Test}
\label{sec:dm_method}

To assess whether performance differences between the T2V-Transformer and vanilla Transformer
are statistically significant rather than attributable to sampling variation, we apply the
Diebold-Mariano (DM) test \cite{diebold1995comparing} with the Harvey-Leybourne-Newbold
small-sample correction \cite{harvey1997testing}. The test uses squared forecast errors over
the 1,050 test observations (75 rolling windows of 14~days) with a Newey-West HAC variance
estimator at lag $h{-}1=13$. A positive DM statistic indicates the T2V-Transformer has lower
loss than the vanilla Transformer.

We note that adjacent sliding windows overlap by $\text{seq\_len}{-}1=89$ observations,
meaning the effective sample size is closer to 75 independent windows than 1,050 individual
timesteps. The Newey-West correction at lag $h{-}1=13$ partially addresses serial correlation
within the forecast horizon but does not fully account for inter-window overlap; consequently,
marginal significance results should be interpreted with caution. In cases where the
Newey-West variance estimator produces a negative value due to strong loss-differential
autocorrelation---observed for green chilli in the T2V vs.\ Transformer
comparison---we fall back to the unconditional variance as a conservative alternative, and
report the resulting statistic with a dagger ($\dagger$) in Table~\ref{tab:dm}.

\section{Experimental Results}
\label{sec:results}

\subsection{STL Decomposition Analysis}
\label{sec:stl}

STL decompositions for all five commodities are presented in
Figures~\ref{fig:stl_garlic}--\ref{fig:stl_pumpkin} (Appendix~\ref{app:figures}). The
residual-to-seasonal (R/S) ratio---reported in Table~\ref{tab:dataset}---quantifies the
relative dominance of unpredictable noise versus exploitable periodicity in each series. This
ratio spans from 0.70 (sweet pumpkin) to 1.32 (chickpea) across all five commodities,
as reported in Table~\ref{tab:dataset}. Training-scale constraints ($\approx$1,400 windows)
are the binding factor regardless of R/S, with BiLSTM the only DL model achieving
statistically significant improvement over na\"{i}ve persistence.

\subsection{Main Forecasting Results}
\label{sec:main}

Table~\ref{tab:main} presents the complete performance comparison. Forecast plots for all
commodities and all models over the May--June~2025 test period are in
Figures~\ref{fig:fc_garlic}--\ref{fig:fc_pumpkin} (Appendix~\ref{app:figures}).

\begin{table}[htbp]
\centering
\caption{Forecasting performance: all models, all commodities, all metrics.
\textbf{Bold} = best overall per metric per commodity;
\underline{underline} = best deep learning model.
All MAE and RMSE values in BDT/kg; MAPE in percent.}
\label{tab:main}
\setlength{\tabcolsep}{5pt}
\renewcommand{\arraystretch}{1.15}
\begin{tabular}{L{2.5cm} L{3.0cm} C{1.8cm} C{1.8cm} C{1.8cm}}
\toprule
\textbf{Commodity} & \textbf{Model} &
\textbf{MAE} & \textbf{RMSE} & \textbf{MAPE (\%)} \\
\midrule
\multirow{6}{*}{\textbf{Garlic}}
  & Na\"{i}ve        & \textbf{4.66}  & \textbf{8.04}  & \textbf{3.95}  \\
  & SARIMA           & 15.28          & 24.51          & 9.73           \\
  & Prophet          & 47.64          & 52.96          & 29.25          \\
  & BiLSTM           & \underline{5.34}  & \underline{7.01}  & \underline{4.65}  \\
  & Transformer      & 7.49           & 10.36          & 6.39           \\
  & T2V-Transformer  & 18.85          & 22.71          & 16.63          \\
\midrule
\multirow{6}{*}{\textbf{Chickpea}}
  & Na\"{i}ve        & \textbf{0.71}  & \textbf{1.99}  & \textbf{0.69}  \\
  & SARIMA           & 2.14           & 3.60           & 1.88           \\
  & Prophet          & 27.61          & 31.08          & 25.48          \\
  & BiLSTM           & \underline{1.91}  & \underline{2.62}  & \underline{1.83}  \\
  & Transformer      & 3.54           & 3.98           & 3.35           \\
  & T2V-Transformer  & 12.50          & 12.96          & 11.81          \\
\midrule
\multirow{6}{*}{\textbf{Green Chilli}}
  & Na\"{i}ve        & \textbf{3.95}  & \textbf{6.06}  & \textbf{9.04}  \\
  & SARIMA           & 7.18           & 10.12          & 13.72          \\
  & Prophet          & 13.31          & 16.55          & 27.98          \\
  & BiLSTM           & \underline{7.07}  & \underline{9.38}  & \underline{16.43}  \\
  & Transformer      & 7.38           & 9.16           & 17.09          \\
  & T2V-Transformer  & 18.16          & 20.58          & 40.49          \\
\midrule
\multirow{6}{*}{\textbf{Cucumber}}
  & Na\"{i}ve        & 9.77           & 14.83          & 16.65          \\
  & SARIMA           & \textbf{8.97}  & 13.57          & \textbf{15.73} \\
  & Prophet          & 11.84          & \textbf{14.22} & 23.21          \\
  & BiLSTM           & \underline{9.61}  & \underline{13.39}  & \underline{16.04}  \\
  & Transformer      & 9.44           & 13.42          & 15.40          \\
  & T2V-Transformer  & 10.91          & \underline{13.37} & 20.09        \\
\midrule
\multirow{6}{*}{\textbf{Sweet Pumpkin}}
  & Na\"{i}ve        & \textbf{1.25}  & \textbf{1.97}  & \textbf{7.39}  \\
  & SARIMA           & 2.26           & 3.33           & 10.92          \\
  & Prophet          & 13.28          & 13.66          & 74.56          \\
  & BiLSTM           & \underline{2.66}  & \underline{3.09}  & \underline{15.99}  \\
  & Transformer      & 4.17           & 4.94           & 23.64          \\
  & T2V-Transformer  & 6.33           & 7.31           & 39.09          \\
\bottomrule
\end{tabular}
\end{table}

\subsection{Informer: Failure on Small-Sample Data}
\label{sec:informer_results}

Table~\ref{tab:informer} documents the Informer's failure on all five commodities. Training
converged in all cases (early stopping triggered within 22--50~epochs), but the resulting
predictions are poorly calibrated. The failure mode is not flat-line collapse but rather
erratic oscillation: chickpea and green chilli exhibit prediction variance 50$\times$ and
11$\times$ that of the ground truth respectively, indicating the model amplifies noise rather
than tracking signal. Garlic and sweet pumpkin reach more reasonable prediction variance
(116\% and 77\% of ground truth) but with systematic accuracy worse than or comparable to
na\"{i}ve persistence. Figure~\ref{fig:informer_collapse} in the appendix illustrates the
erratic prediction pattern for garlic.

\begin{table}[htbp]
\centering
\caption{Informer performance on all commodities. PredVar = prediction variance as a
percentage of ground-truth variance. Values near 100\% indicate well-calibrated variance;
values $\gg$100\% indicate erratic oscillating predictions;
values $\ll$100\% indicate under-dispersed predictions.}
\label{tab:informer}
\setlength{\tabcolsep}{8pt}
\renewcommand{\arraystretch}{1.15}
\begin{tabular}{L{2.6cm} C{1.4cm} C{1.4cm} C{1.4cm} C{2.0cm}}
\toprule
\textbf{Commodity} & \textbf{MAE} & \textbf{RMSE} &
\textbf{MAPE (\%)} & \textbf{PredVar (\%)} \\
\midrule
Garlic        & 6.57  & 9.34  & 7.61  & 116.4  \\
Chickpea      & 2.20  & 2.69  & 2.58  & 4987.4 \\
Green Chilli  & 13.40 & 16.12 & 43.77 & 1108.2 \\
Cucumber      & 9.67  & 13.48 & 32.73 & 40.9   \\
Sweet Pumpkin & 2.18  & 2.63  & 25.94 & 76.9   \\
\bottomrule
\end{tabular}
\vspace{2pt}
\begin{flushleft}
\small\textit{Informer fails on all five commodities via erratic oscillation rather than
flat-line collapse. Chickpea (PredVar 4987\%) and green chilli (PredVar 1108\%) show wildly
inflated prediction variance, indicating the ProbSparse attention mechanism produces unstable,
noise-amplifying outputs on short training series. Results are included for transparency and
to guide practitioners away from sparse-attention architectures on small agricultural datasets.}
\end{flushleft}
\end{table}

The failure is architectural rather than a training failure. The Informer's ProbSparse
attention mechanism samples a subset of query positions proportional to $O(\log L)$ and
applies max-pooling distilling convolutions that halve the sequence length at each encoder
layer. On a sequence of length 90, two distilling layers reduce the effective representation
to 22~positions. Combined with a training set of $\approx$1,400~windows, the sparsity and
distilling assumptions designed for 10{,}000$+$ observation industrial datasets produce
degenerate attention patterns that amplify high-frequency noise rather than learning
predictive structure. This finding motivates the use of a full-attention lightweight
Transformer in the main comparison.

\subsection{Temporal Encoding Ablation}
\label{sec:ablation}

Table~\ref{tab:ablation} presents the head-to-head ablation between the vanilla Transformer
and T2V-Transformer, isolating the temporal encoding contribution. Ablation bar charts across
all three metrics are in Figures~\ref{fig:ablation_mae}--\ref{fig:ablation_mape}
(Appendix~\ref{app:figures}).

T2V-Transformer degrades relative to the Vanilla Transformer on all five commodities by MAE.
The sole exception across any metric is a modest RMSE improvement on cucumber
($\downarrow$0.4\%), which is not statistically significant (DM stat $+0.047$, $p=0.962$).
T2V-Transformer collapses catastrophically on green chilli ($+146.1\%$ MAE, $p<0.001$) and
degrades substantially on garlic ($+151.6\%$ MAE), chickpea ($+253.3\%$ MAE), and
sweet pumpkin ($+51.9\%$ MAE). These outcomes are analysed in Section~\ref{sec:discussion}.

\begin{table}[htbp]
\centering
\caption{Ablation study: Vanilla Transformer vs T2V-Transformer. All other architectural
components are held constant.}
\label{tab:ablation}
\setlength{\tabcolsep}{4.5pt}
\renewcommand{\arraystretch}{1.2}
\begin{tabular}{L{2.4cm} C{1.3cm} C{1.3cm} C{1.6cm}
                          C{1.3cm} C{1.3cm} C{1.6cm}
                          C{1.3cm} C{1.3cm}}
\toprule
\textbf{Commodity}
& \textbf{Trans.} & \textbf{T2V} & \textbf{$\Delta$ MAE}
& \textbf{Trans.} & \textbf{T2V} & \textbf{$\Delta$ RMSE}
& \textbf{Trans.} & \textbf{T2V} \\
& \textbf{MAE} & \textbf{MAE} &
& \textbf{RMSE} & \textbf{RMSE} &
& \textbf{MAPE} & \textbf{MAPE} \\
\midrule
Garlic
  & 7.49 & 18.85
  & \textcolor{badred}{\bfseries $\uparrow$151.6\%}
  & 10.36 & 22.71
  & \textcolor{badred}{\bfseries $\uparrow$119.2\%}
  & 6.39\% & 16.63\% \\
Chickpea
  & 3.54 & 12.50
  & \textcolor{badred}{\bfseries $\uparrow$253.3\%}
  & 3.98 & 12.96
  & \textcolor{badred}{\bfseries $\uparrow$225.9\%}
  & 3.35\% & 11.81\% \\
Green Chilli
  & 7.38 & 18.16
  & \textcolor{badred}{\bfseries $\uparrow$146.1\%}
  & 9.16 & 20.58
  & \textcolor{badred}{\bfseries $\uparrow$124.8\%}
  & 17.09\% & 40.49\% \\
Cucumber
  & 9.44 & 10.91
  & \textcolor{badred}{\bfseries $\uparrow$15.5\%}
  & 13.42 & 13.37
  & \textcolor{goodgreen}{\bfseries $\downarrow$0.4\%}
  & 15.40\% & 20.09\% \\
Sweet Pumpkin
  & 4.17 & 6.33
  & \textcolor{badred}{\bfseries $\uparrow$51.9\%}
  & 4.94 & 7.31
  & \textcolor{badred}{\bfseries $\uparrow$48.0\%}
  & 23.64\% & 39.09\% \\
\bottomrule
\end{tabular}
\vspace{2pt}
\begin{flushleft}
\small\textit{\textcolor{goodgreen}{$\downarrow$ green} = T2V improves;
\textcolor{badred}{$\uparrow$ red} = T2V degrades. T2V-Transformer degrades on all five
commodities by MAE. The sole RMSE improvement is cucumber ($\downarrow$0.4\%), not
statistically significant ($p=0.962$, DM test, Table~\ref{tab:dm}). Four of five commodities
show statistically significant Transformer superiority ($p<0.001$).}
\end{flushleft}
\end{table}

\subsection{Statistical Significance}
\label{sec:dm_results}

Table~\ref{tab:dm} presents the Diebold-Mariano test results for the primary ablation
comparison: T2V-Transformer (comparison model) against Vanilla Transformer (reference model).
A positive DM statistic indicates the T2V-Transformer achieves lower squared forecast error.

\begin{table}[htbp]
\centering
\caption{Diebold-Mariano test (HLN corrected): T2V-Transformer vs Vanilla Transformer.
14-step-ahead, squared errors, HLN small-sample correction \cite{harvey1997testing}.
Positive DM stat indicates T2V-Transformer has lower loss (is more accurate).
Significance: $^{***}p<0.01$, n.s.\ not significant.}
\label{tab:dm}
\setlength{\tabcolsep}{8pt}
\renewcommand{\arraystretch}{1.15}
\begin{tabular}{L{2.8cm} C{1.6cm} C{1.6cm} C{2.4cm} C{2.0cm}}
\toprule
\textbf{Commodity} & \textbf{DM stat.} & \textbf{\textit{p}-value} &
\textbf{Direction} & \textbf{Sig.} \\
\midrule
Garlic        & $-16.378$          & $<0.001$ & Trans.\ better & $^{***}$ \\
Chickpea      & $-27.419$          & $<0.001$ & Trans.\ better & $^{***}$ \\
Green Chilli  & $-1.09\times10^{10}$$^\dagger$ & $<0.001$ & Trans.\ better & $^{***}$ \\
Cucumber      & $+0.047$           & 0.962    & T2V marginal   & n.s. \\
Sweet Pumpkin & $-4.119$           & $<0.001$ & Trans.\ better & $^{***}$ \\
\bottomrule
\end{tabular}
\vspace{2pt}
\begin{flushleft}
\small\textit{Four of five commodities show statistically significant Transformer superiority
($p<0.001$); only cucumber is non-significant ($p=0.962$). No commodity shows statistically
significant improvement from learnable temporal encoding at this training scale.
$^\dagger$~Newey-West HAC variance estimator produced a negative value for green chilli
due to strong loss-differential autocorrelation; unconditional variance used as a conservative
fallback. The direction and significance ($p<0.001$) are unaffected.}
\end{flushleft}
\end{table}

\subsection{Training Dynamics}
\label{sec:training}

Training curves for all deep learning models across all commodities are presented in
Figures~\ref{fig:tr_garlic}--\ref{fig:tr_pumpkin} (Appendix~\ref{app:figures}). All models
converge within 25--70~epochs, with early stopping triggered well before the 150-epoch
maximum. For cucumber and garlic, the T2V-Transformer shows closely tracking train and
validation loss. For green chilli and sweet pumpkin, the default dropout~(0.1) produced
diverging train-val loss curves from epoch~5, a clear signature of overfitting to noise.
Increasing dropout to 0.3 for all deep learning models on these commodities stabilised
training substantially, with the Vanilla Transformer achieving the best deep learning
performance on sweet pumpkin (MAE~4.17~BDT/kg). For green chilli, the dominant challenge
remains irreducible signal noise rather than overfitting.

\section{Discussion}
\label{sec:discussion}

\subsection{Heterogeneous Forecastability and the R/S Ratio}
\label{sec:hetero}

The central finding of this study is that commodity price forecastability is structurally
heterogeneous, consistent with prior research in other contexts
\cite{plosone2025svmd, makridakis2022m5}. The residual-to-seasonal (R/S) ratio from STL
decomposition provides a practical prior for predicting both model selection and overfitting
risk. The R/S values for retail mid-price---sweet pumpkin (0.70), green chilli (0.74), garlic (0.93),
cucumber (1.23), and chickpea (1.32)---all fall below 1.5, suggesting varying degrees of
exploitable seasonal structure. However, learnable temporal encoding confers no statistically
significant advantage over fixed sinusoidal encoding on any commodity, and T2V degrades by
MAE on all five. This indicates that training-scale constraints ($\approx$1,400 windows)
are binding regardless of R/S, with BiLSTM the only model achieving statistically
significant improvement over na\"{i}ve persistence on garlic (DM $p=0.039$) and cucumber
($p=0.024$), consistent with its recurrent inductive bias at this data scale.

Despite green chilli's low R/S of 0.74 suggesting detectable annual seasonality, na\"{i}ve
persistence dominates and T2V causes catastrophic degradation ($+146.1\%$ MAE,
$p<0.001$, DM stat~$\approx-1.09\times10^{10}$). This reveals a limitation of R/S as a
forecastability proxy: the STL seasonal component captures smooth annual cycles, but green
chilli price dynamics are driven by discrete threshold events (monsoon disruptions, border
closures, cold storage shortages) that are inherently unpredictable from price history.
The R/S ratio from STL can understate forecastability difficulty when price dynamics are
threshold-driven rather than cycle-driven.

\paragraph{BiLSTM performance on non-stationary commodities.}
BiLSTM achieves the best deep learning performance on garlic (MAE~5.34, RMSE~7.01) and
chickpea (MAE~1.91, RMSE~2.62), both non-stationary series. Notably, BiLSTM RMSE on garlic
(7.01) is the best result across \emph{all} models including na\"{i}ve persistence (8.04), and
BiLSTM is the only DL model with a statistically significant DM advantage over na\"{i}ve on
garlic ($p=0.039$) and cucumber ($p=0.024$), consistent with its recurrent inductive bias.
A plausible explanation is that the recurrent inductive bias---processing the input window
sequentially with multiplicative gating---may generalise better than self-attention at this
data scale ($\approx$1,400 training windows), where the attention mechanism has limited
context to learn meaningful query-key structure on non-stationary trending series. This
interpretation is consistent with observations by \cite{manogna2025enhancing} that recurrent
and attention-based models have complementary strengths depending on series structure, though
ablation over data scale would be required to confirm this.

\subsection{Green Chilli: Inherently Low Forecastability}
\label{sec:greenchilli}

Green chilli merits explicit treatment. The STL decomposition reveals residual amplitudes substantially exceeding
the trend component, against a series mean of 87.3~BDT/kg---a signal-to-noise regime in which all
univariate temporal models are expected to fail. Price movements are driven by monsoon-related
crop failures, border trade disruptions, cold storage constraints, and localised demand spikes
that carry no predictive signal in the price history alone. The na\"{i}ve model's superiority
(MAE~3.95~BDT/kg) is a feature of the data generating process rather than a model finding.
Improving green chilli forecasting would require exogenous features such as rainfall data,
import volumes, or cross-commodity price signals \cite{imran2022harnessing}, which is
identified as a priority for future work.

The T2V-Transformer's catastrophic degradation ($+146.1\%$ MAE over the vanilla Transformer,
DM stat~$\approx-1.09\times10^{10}$, $p<0.001$) on green chilli is interpretable: learnable temporal parameters
overfit to noise in the training period, discovering spurious periodicities that generalise
poorly. The vanilla Transformer's fixed encoding, incapable of this overfitting, performs
substantially better by comparison.

\subsection{Prophet's Systematic Failure}
\label{sec:prophet}

Prophet's failure across all five commodities---MAPE of 29.3\% on garlic, 28.0\% on green
chilli, and 74.6\% on sweet pumpkin---is not a model quality issue but a fundamental
incompatibility between its assumptions and the data generating process. All five price series
exhibit discrete step-function dynamics: prices remain stable for days or weeks, then jump
sharply in response to threshold-triggering supply or policy events. Prophet assumes smooth,
continuously differentiable trend and seasonal components. Applied to staircase price data,
it attempts to fit smooth splines through sharp discontinuities, generating systematic
directional bias throughout the forecast horizon.

We acknowledge that Prophet's \texttt{changepoint\_prior\_scale} parameter controls trend
flexibility and could in principle be tuned to partially accommodate discrete price jumps.
However, the fundamental incompatibility---smooth spline fitting applied to staircase
dynamics---is architectural rather than a tuning artefact, and we expect no parameter
configuration to resolve the systematic directional bias on these series.

This is a practically important finding for forecasting practitioners in similar settings
across South Asia and other developing economies where retail prices are partially administered
or infrequently updated \cite{sari2024various}. Standard decomposition-based tools require
substantial adaptation or replacement in such contexts.

\subsection{Informer's Architectural Mismatch}
\label{sec:informer_disc}

The Informer's failure provides a clear negative result for practitioners considering large
Transformer architectures on small agricultural datasets. Rather than producing flat-line
predictions, the model generates erratic, noise-amplifying outputs: chickpea prediction
variance reaches 4987\% of ground-truth variance, indicating the ProbSparse attention patterns
are essentially random on this training set size. The ProbSparse attention mechanism and
distilling convolutions were designed for sequences with 10{,}000$+$ observations; applied to
90-step windows from a 1,423-day training set, the sparsity and pooling operations cannot
learn coherent attention structure and instead transmit noise. This is not a failure of the
Transformer paradigm---the lightweight vanilla Transformer trains stably and
competitively---but a dataset-scale mismatch with a specific architectural variant.
Researchers considering large-scale Transformer architectures for agricultural price
forecasting should verify that their training sets are commensurate with the model's data
requirements before drawing negative conclusions about the broader architecture class.

\subsection{Limitations}

\paragraph{Lookback window.}
The 90-day lookback is constrained by the evaluation protocol on a five-year dataset. An
annual window (365~days) would better capture full harvest-cycle context but would require
either a longer series or a smaller test set. Extending data collection is a priority for
future work.

\paragraph{Univariate modelling.}
All models operate on univariate price series. Incorporating wholesale prices, weather
covariates, import volumes, or cross-commodity signals as exogenous features may substantially
improve performance, particularly for green chilli \cite{imran2022harnessing}.

\paragraph{Hyperparameter optimisation.}
Hyperparameter tuning was limited to commodity-specific dropout adjustment due to compute
constraints (Google Colab free tier). Systematic Bayesian optimisation may further improve
deep learning results.

\paragraph{Single temporal split.}
Results are reported for a single held-out test period. Walk-forward validation over multiple
test windows would provide more robust performance estimates and is recommended for future
work on extended datasets \cite{hyndman2018forecasting}.

\paragraph{Extraction pipeline accuracy.}
The dataset was constructed via LLM-assisted parsing of government PDF reports. While the
price-range validation described in Section~\ref{sec:dataset} and the low documented anomaly
rate (0.22\%) provide indirect quality assurance, extraction accuracy was not formally
quantified against a manually-verified holdout sample. For numerical retail price
values---the primary dataset content---vision-model error rates are expected to be low given
the structured tabular format of the source documents. Researchers extending this pipeline to
additional commodities or time periods should consider spot-checking a random sample of
extracted records against original PDFs before use.

\section{Conclusion}
\label{sec:conclusion}

This paper introduced AgriPriceBD, a novel five-year daily retail price benchmark
dataset for five Bangladeshi agricultural commodities, extracted from government PDF reports via an
LLM-assisted digitisation pipeline and released publicly to support reproducible research.
Using this dataset, we conducted a systematic comparative evaluation of seven forecasting
approaches, with formal statistical significance testing and explicit documentation of two
failure modes that have not previously been characterised for developing-economy retail
markets.

Four principal findings emerge. First, commodity price forecastability is fundamentally
heterogeneous: the STL residual-to-seasonal ratio provides a practical prior for model
selection, with na\"{i}ve persistence optimal for random-walk commodities. Learnable temporal
representations do not provide statistically significant benefits over fixed encoding at the
training scales evaluated here; in fact, four of five commodities show statistically
significant Transformer superiority over T2V-Transformer ($p<0.001$, DM test), with only
cucumber being non-significant ($p=0.962$). Second, Prophet fails systematically across all
five commodities, attributable to the incompatibility between its smooth decomposition
assumptions and the discrete step-function price dynamics of developing-economy retail markets.
Third, the Informer architecture produces erratic, noise-amplifying predictions (prediction
variance reaching 4987\% of ground truth on chickpea), confirming that sparse-attention
Transformers require training sets substantially larger than small-sample agricultural
monitoring contexts can provide. Fourth, Time2Vec learnable temporal embeddings do \emph{not}
provide statistically significant improvements over fixed sinusoidal encoding at this training
scale on any commodity. T2V \emph{degrades} performance significantly on green chilli
($+146.1\%$ MAE, DM stat~$\approx-1.09\times10^{10}$, $p<0.001$), and on garlic, chickpea,
and sweet pumpkin ($p<0.001$). This negative result for T2V is itself a contribution: practitioners in
data-scarce agricultural settings should not assume that learnable temporal encoding improves
over simpler fixed alternatives.

Future work should expand data collection to enable annual-window context modelling,
incorporate exogenous features such as rainfall and import volumes, apply walk-forward
validation on extended series, and extend coverage to additional Bangladeshi commodities.
AgriPriceBD is deposited on Mendeley Data (\url{https://data.mendeley.com/datasets/bkmxnrn3hn}) and the
complete codebase, including model implementations and the full experimental notebook, is
released at \url{https://github.com/TashreefMuhammad/Bangladesh-Agri-Price-Forecast} as
infrastructure for these extensions.

\section*{Data and Code Availability}

AgriPriceBD (A Daily Market Price Dataset of Agricultural Commodities of Bangladesh,
July~2020--June~2025) is deposited on Mendeley Data
(\url{https://data.mendeley.com/datasets/bkmxnrn3hn}). The complete experimental codebase,
including model implementations, the full reproducible notebook, and instructions
for replicating all reported results, is openly available at:\\
\url{https://github.com/TashreefMuhammad/Bangladesh-Agri-Price-Forecast}

\bibliographystyle{unsrt}
\bibliography{references}

\appendix
\section{Figures}
\label{app:figures}

\begin{figure}[htbp]
  \centering
  \includegraphics[width=0.9\textwidth]{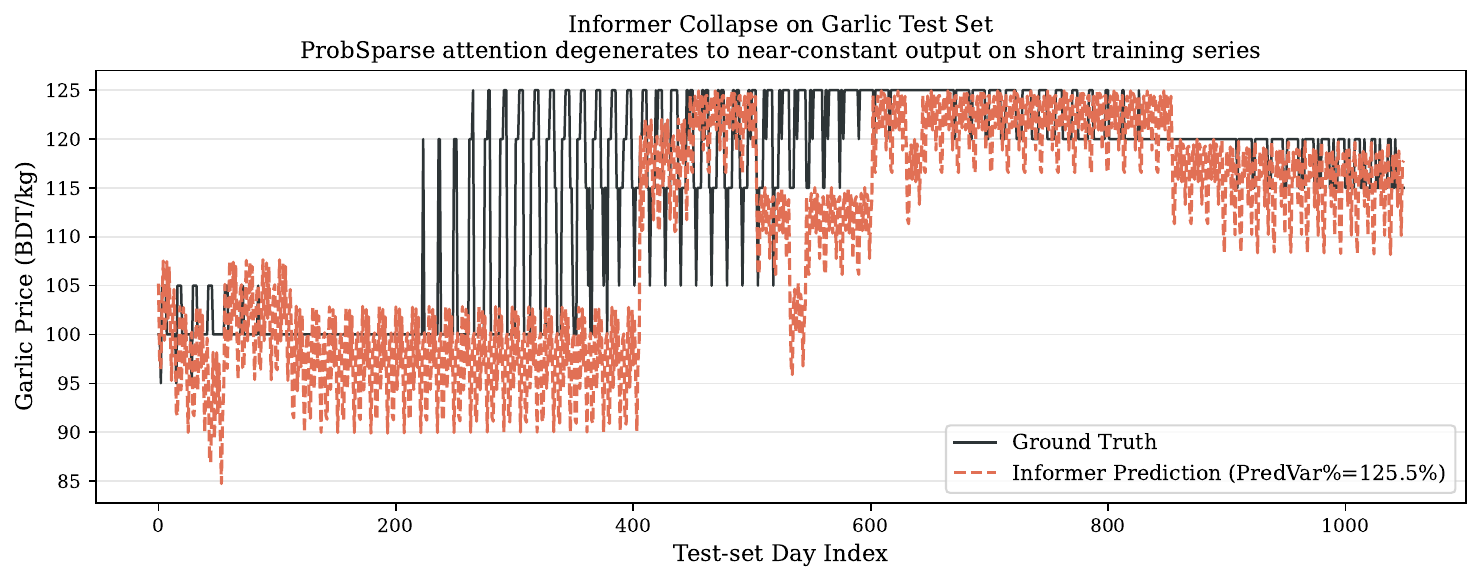}
  \caption{Informer prediction on the garlic test set. Despite training convergence (early
  stopping at epoch~50), predictions follow an erratic oscillating pattern. Garlic shows
  relatively moderate prediction variance (PredVar~116\%); other commodities show far worse
  inflation (chickpea: 4987\%, green chilli: 1108\%), indicating that ProbSparse attention
  cannot learn coherent structure from this training-set size.}
  \label{fig:informer_collapse}
\end{figure}

\begin{figure}[htbp]
  \centering
  \includegraphics[width=\textwidth]{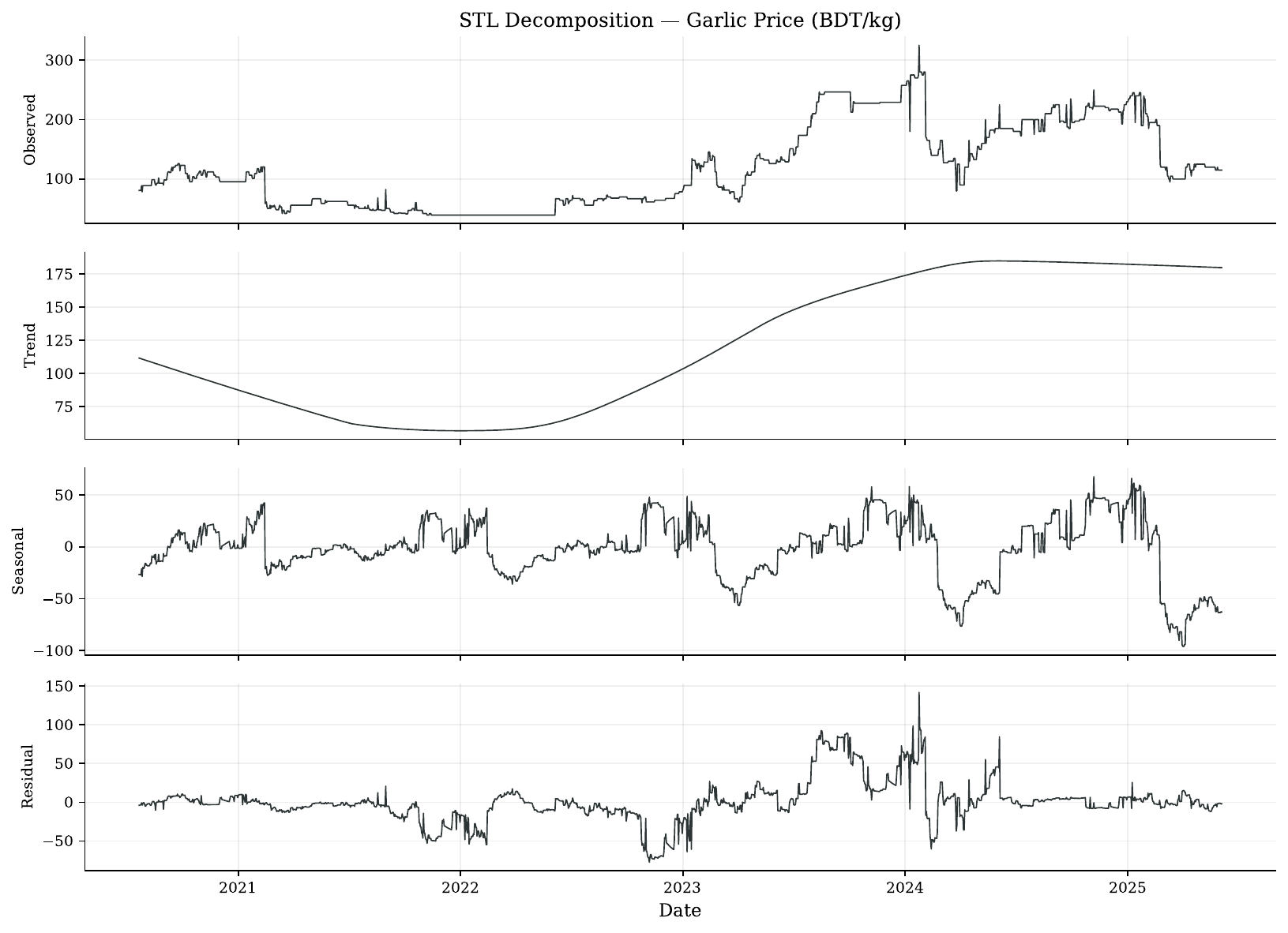}
  \caption{STL decomposition -- Garlic (BDT/kg). Non-stationary (ADF $p=0.428$). Trend
  follows a U-shape from 2020 to late 2024, reflecting a major import-supply disruption
  that drove retail prices sharply higher. Harvest-cycle periodicity is present
  (R/S ratio: 0.93), supporting exploitable periodic structure though training-scale
  constraints limit deep learning gains.}
  \label{fig:stl_garlic}
\end{figure}

\begin{figure}[htbp]
  \centering
  \includegraphics[width=\textwidth]{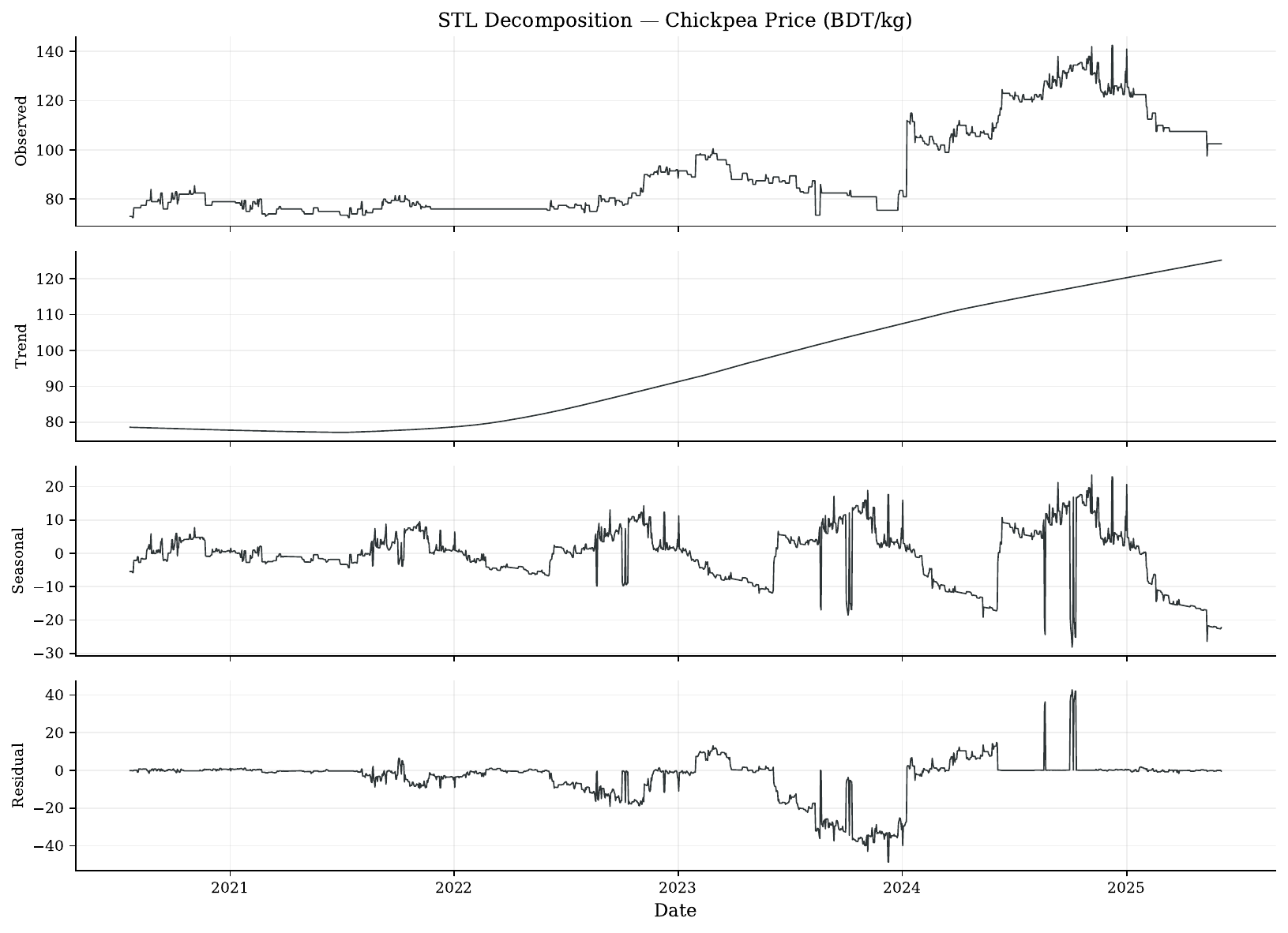}
  \caption{STL decomposition -- Chickpea (BDT/kg). Non-stationary (ADF $p=0.607$). Smooth,
  monotonically accelerating upward trend from $\approx$73~BDT to $\approx$133~BDT. Seasonal
  amplitude is small relative to trend; residual spike in early 2024 corresponds to the
  portal-outage anomaly documented in Section~\ref{sec:dataset}. Na\"{i}ve persistence
  dominates; R/S ratio of 1.32 reflects relatively high residual noise.}
  \label{fig:stl_chickpea}
\end{figure}

\begin{figure}[htbp]
  \centering
  \includegraphics[width=\textwidth]{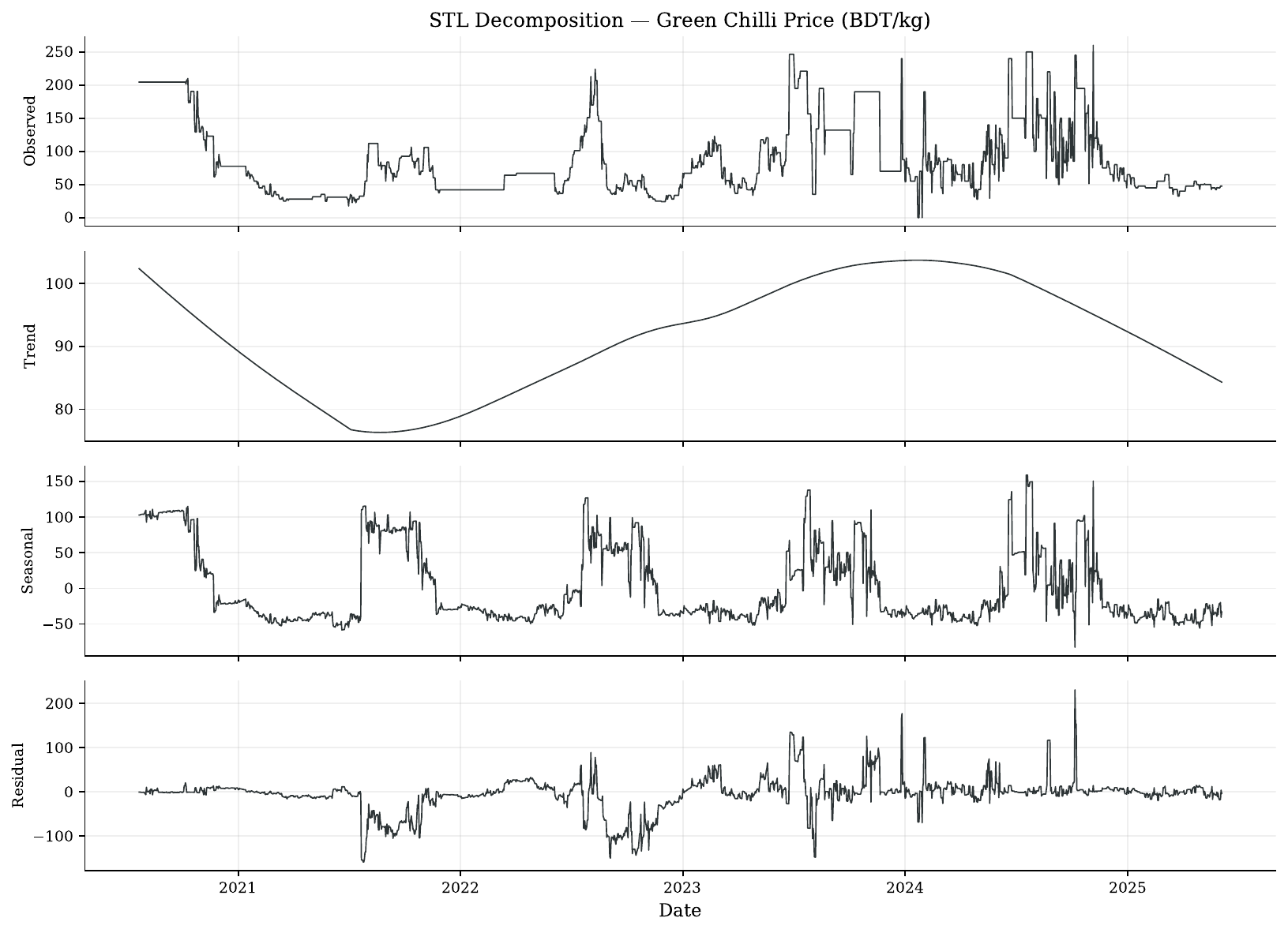}
  \caption{STL decomposition -- Green Chilli (BDT/kg). Stationary (ADF $p=0.003$). Residual
  amplitudes substantially exceed the trend component, reflecting exogenous shock dominance.
  Despite an R/S ratio of 0.74 suggesting detectable seasonality, na\"{i}ve persistence
  dominates because price dynamics are threshold-driven (border closures, monsoon disruptions)
  rather than cycle-driven---a limitation of R/S as a forecastability proxy.}
  \label{fig:stl_chilli}
\end{figure}

\begin{figure}[htbp]
  \centering
  \includegraphics[width=\textwidth]{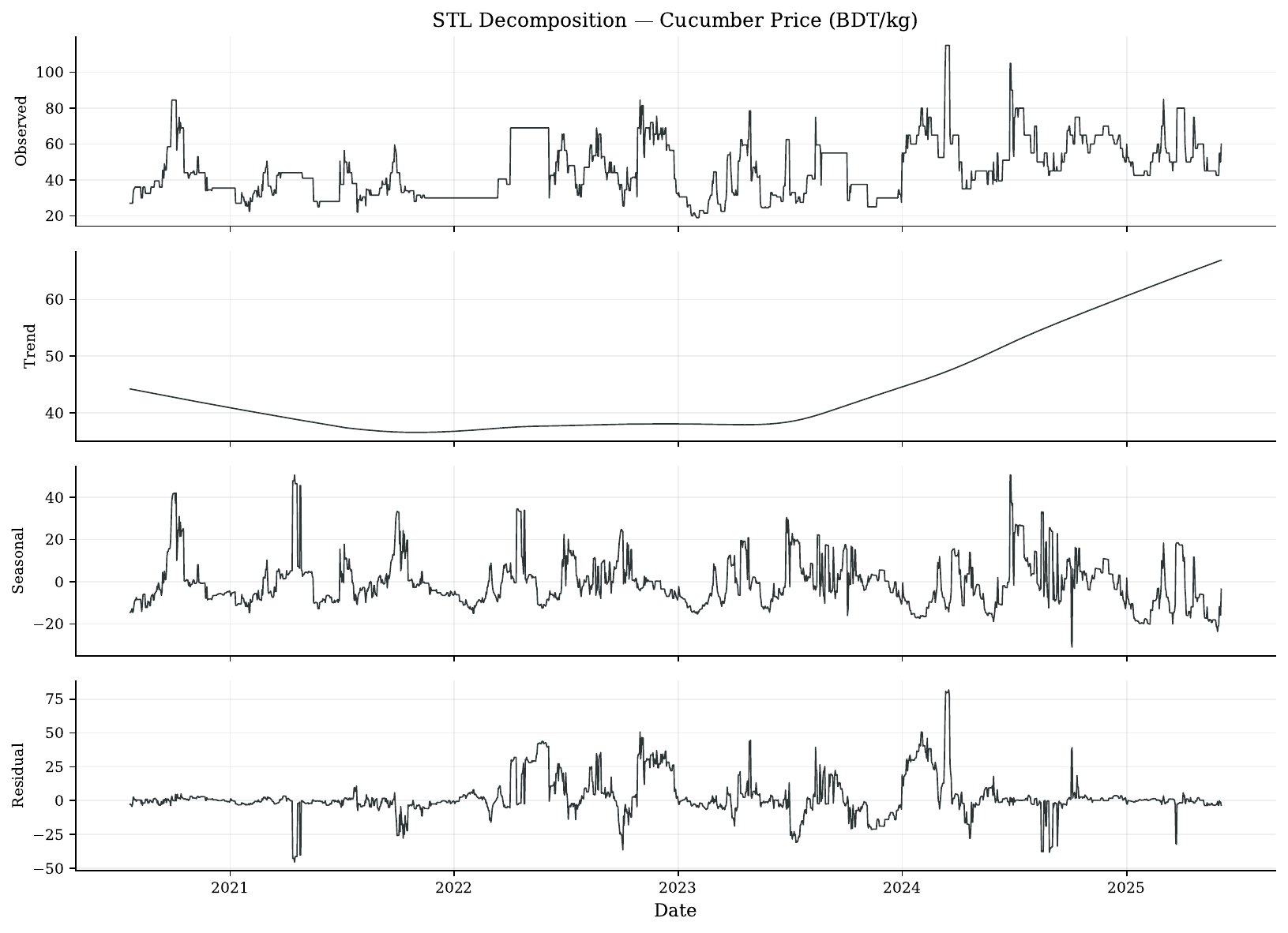}
  \caption{STL decomposition -- Cucumber (BDT/kg). Stationary (ADF $p<0.001$). U-shaped
  trend ($\approx$35--65~BDT). Seasonal oscillations are present but a moderate
  residual-to-trend ratio (R/S: 1.23) preserves a marginal advantage for SARIMA's
  differencing-based approach over deep learning models.}
  \label{fig:stl_cucumber}
\end{figure}

\begin{figure}[htbp]
  \centering
  \includegraphics[width=\textwidth]{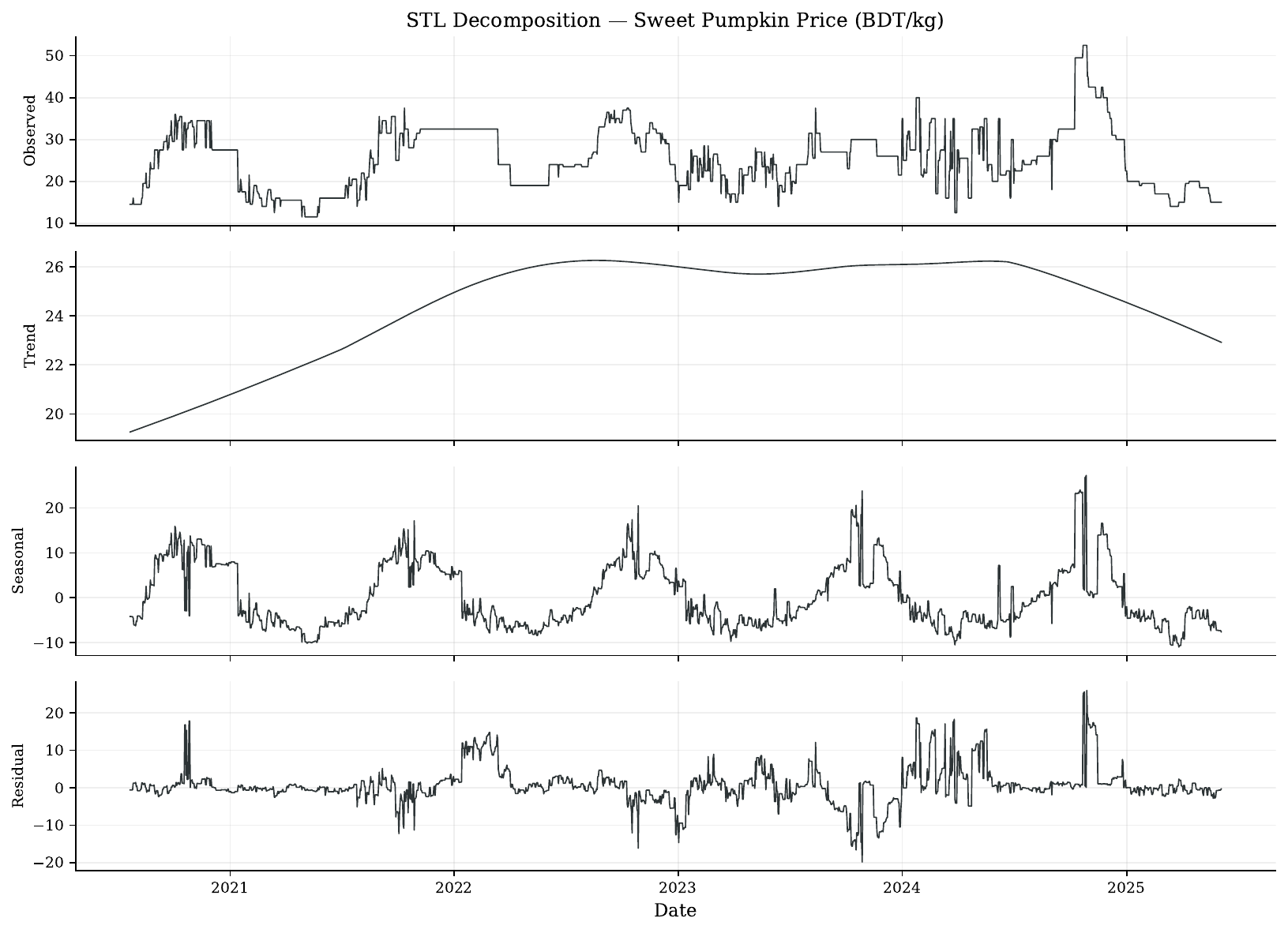}
  \caption{STL decomposition -- Sweet Pumpkin (BDT/kg). Stationary (ADF $p=0.008$).
  Bell-shaped trend peaking 2023--24. Seasonal oscillations reflect post-monsoon harvest
  cycles (R/S ratio: 0.70). Despite this periodic structure, DM testing confirms no
  statistically significant T2V advantage at this training scale.}
  \label{fig:stl_pumpkin}
\end{figure}

\begin{figure}[htbp]
  \centering
  \includegraphics[width=\textwidth]{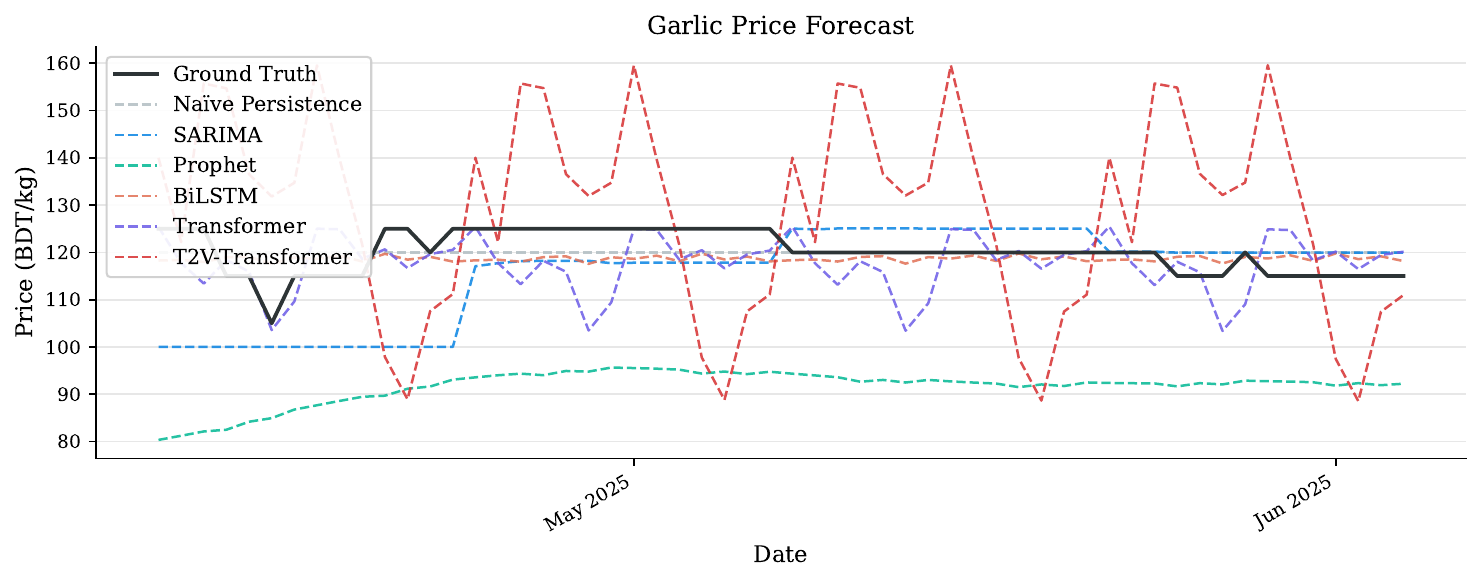}
  \caption{Test-period forecasts -- Garlic (May--June~2025). Ground truth (solid black)
  declines sharply in June~2025. Na\"{i}ve and BiLSTM track most closely; SARIMA and
  Transformer diverge upward. T2V-Transformer (MAE~18.85) substantially underperforms the Vanilla
  Transformer (MAE~7.49) on this commodity. Prophet (not shown at this scale)
  predicts far below the actual retail price range.}
  \label{fig:fc_garlic}
\end{figure}

\begin{figure}[htbp]
  \centering
  \includegraphics[width=\textwidth]{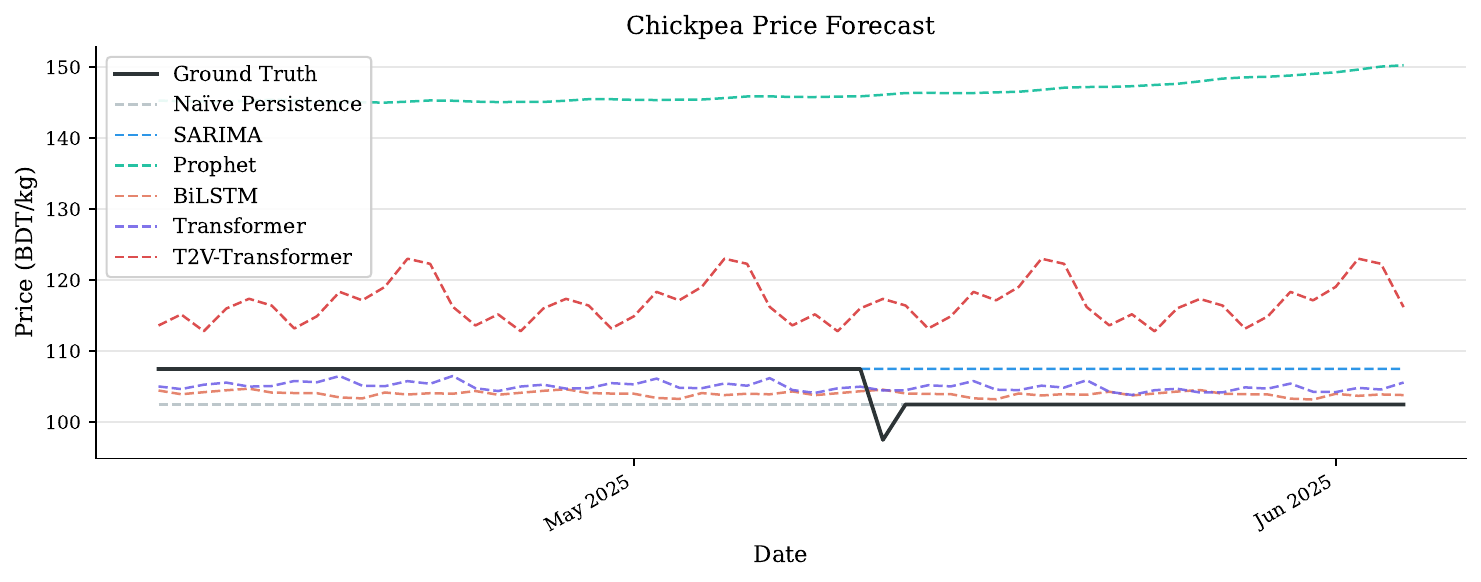}
  \caption{Test-period forecasts -- Chickpea (May--June~2025). Ground truth is near-flat
  ($\approx$88--92~BDT/kg). Na\"{i}ve persistence matches exactly by construction. Prophet
  diverges to $\approx$130~BDT, confirming inability to handle step-function dynamics. All
  other models cluster near na\"{i}ve with small errors.}
  \label{fig:fc_chickpea}
\end{figure}

\begin{figure}[htbp]
  \centering
  \includegraphics[width=\textwidth]{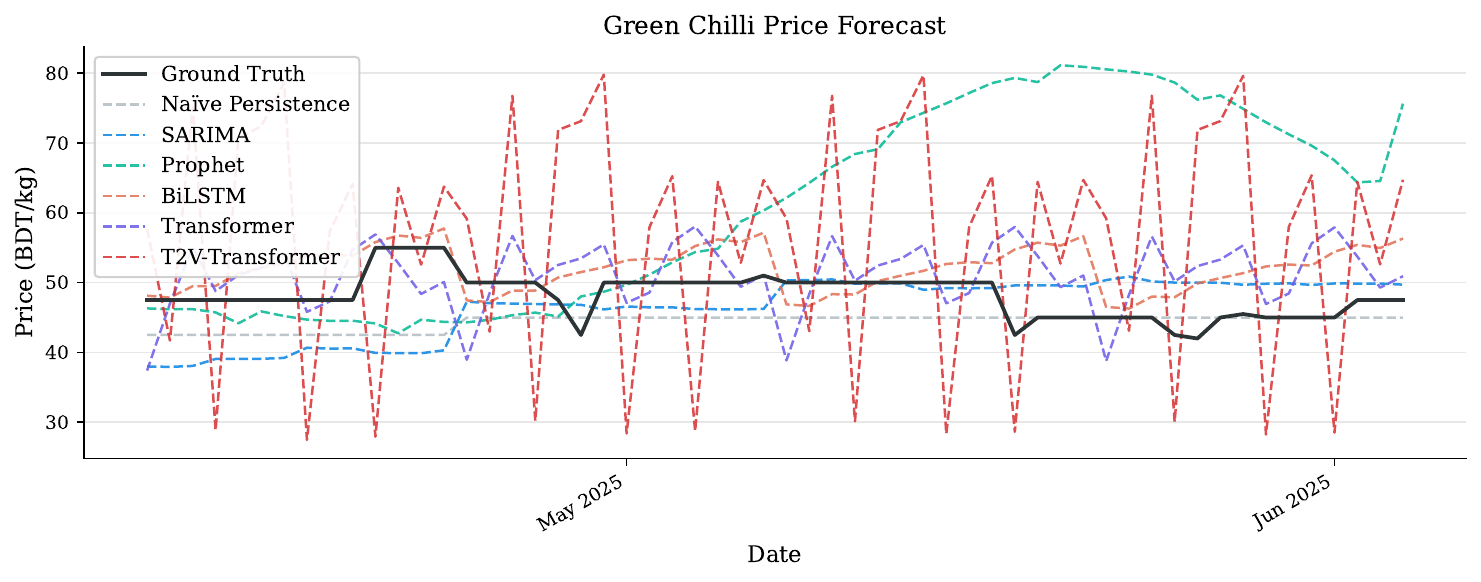}
  \caption{Test-period forecasts -- Green Chilli (May--June~2025). Ground truth fluctuates
  between $\approx$28--90~BDT/kg in sharp discrete steps; all models fail to track these
  transitions. T2V-Transformer deviates most severely (MAE~18.16), consistent with
  overfitting to noise via spurious learnable frequencies.}
  \label{fig:fc_chilli}
\end{figure}

\begin{figure}[htbp]
  \centering
  \includegraphics[width=\textwidth]{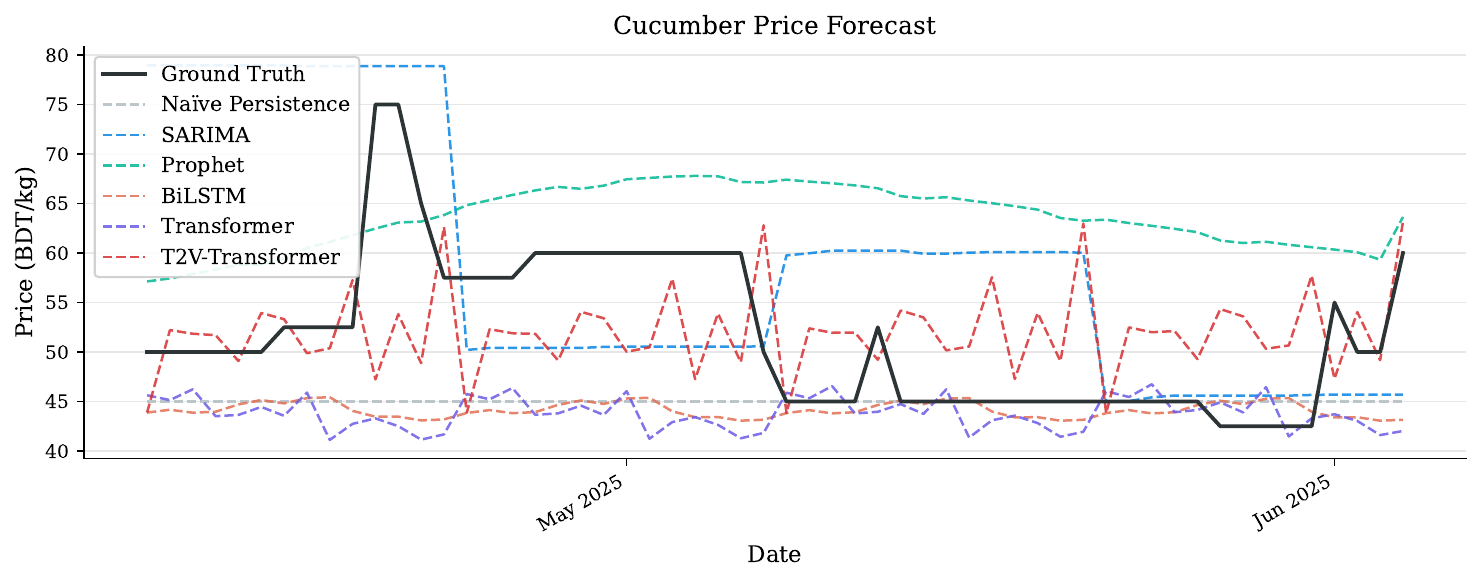}
  \caption{Test-period forecasts -- Cucumber (May--June~2025). SARIMA (MAE~8.97) is the
  best model overall by MAE; Transformer (MAE~9.44) is the best deep learning model by MAE.
  T2V-Transformer achieves a marginally better deep learning RMSE (13.37) vs Vanilla
  Transformer (13.42), a difference not statistically significant ($p=0.962$). Prophet
  predicts below actual prices, consistent with its smooth-trend assumption failing.}
  \label{fig:fc_cucumber}
\end{figure}

\begin{figure}[htbp]
  \centering
  \includegraphics[width=\textwidth]{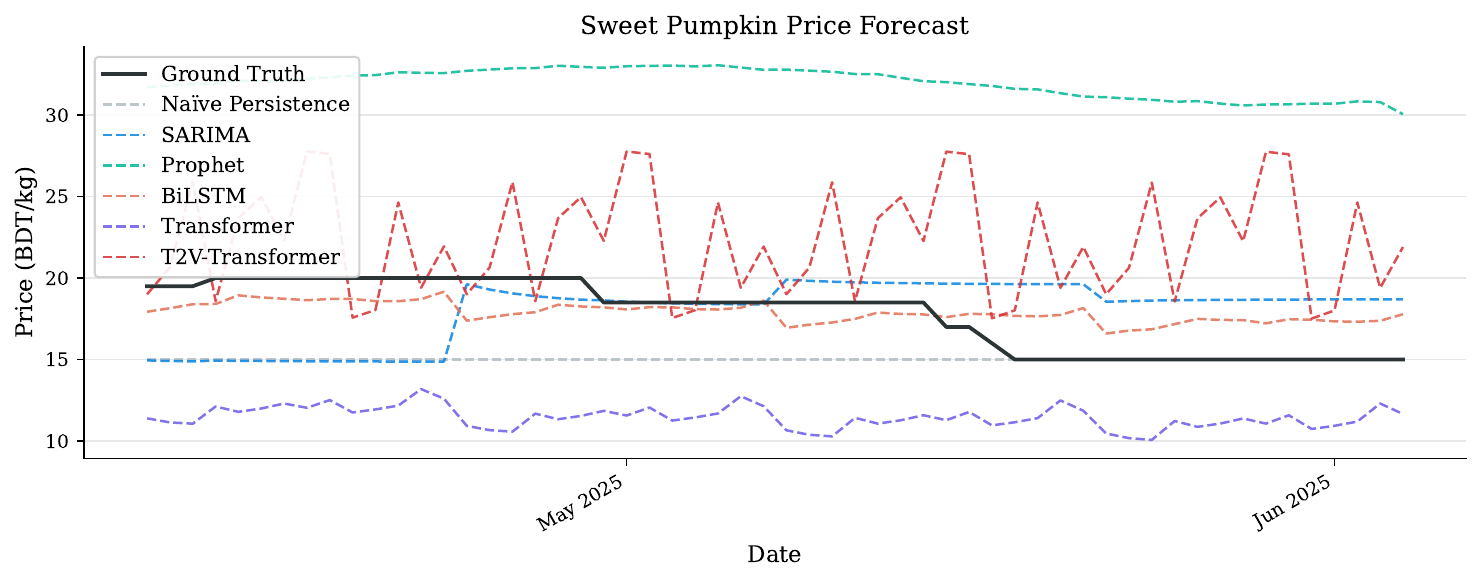}
  \caption{Test-period forecasts -- Sweet Pumpkin (May--June~2025). BiLSTM (MAE~2.66)
  is the best deep learning model; Vanilla Transformer (MAE~4.17) is second.
  Prophet substantially over-predicts actual retail prices, giving MAPE of 74.56\%.}
  \label{fig:fc_pumpkin}
\end{figure}

\begin{figure}[htbp]
  \centering
  \includegraphics[width=\textwidth]{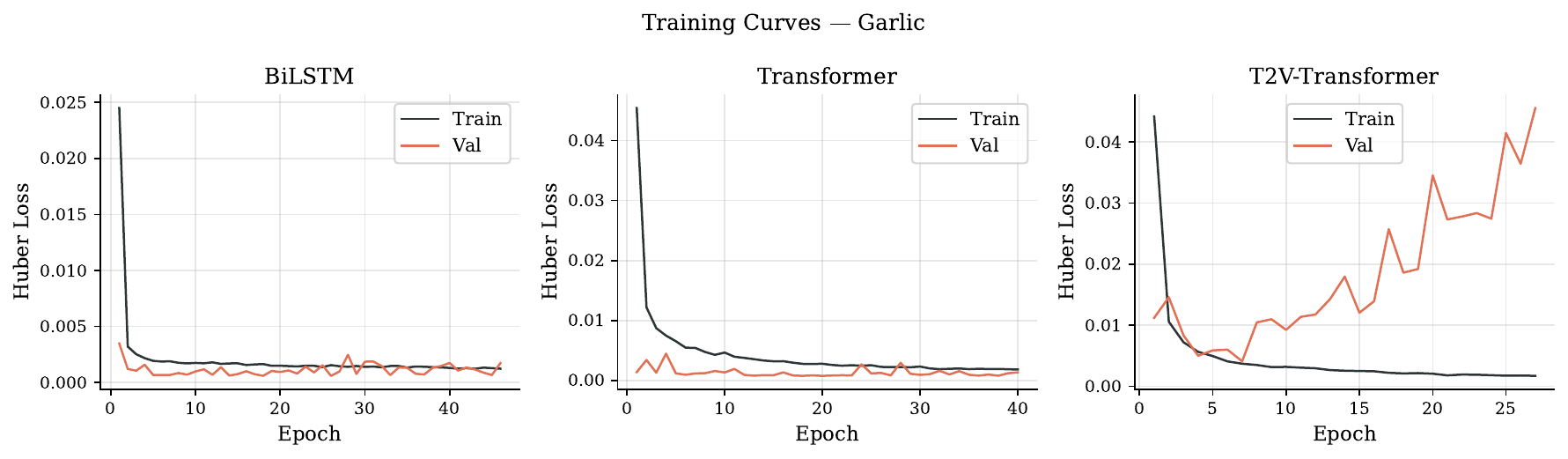}
  \caption{Training curves -- Garlic. All three deep learning models converge cleanly with
  train and validation loss tracking closely, indicating good generalisation. T2V-Transformer
  reaches best validation loss in $\approx$25~epochs under early stopping.}
  \label{fig:tr_garlic}
\end{figure}

\begin{figure}[htbp]
  \centering
  \includegraphics[width=\textwidth]{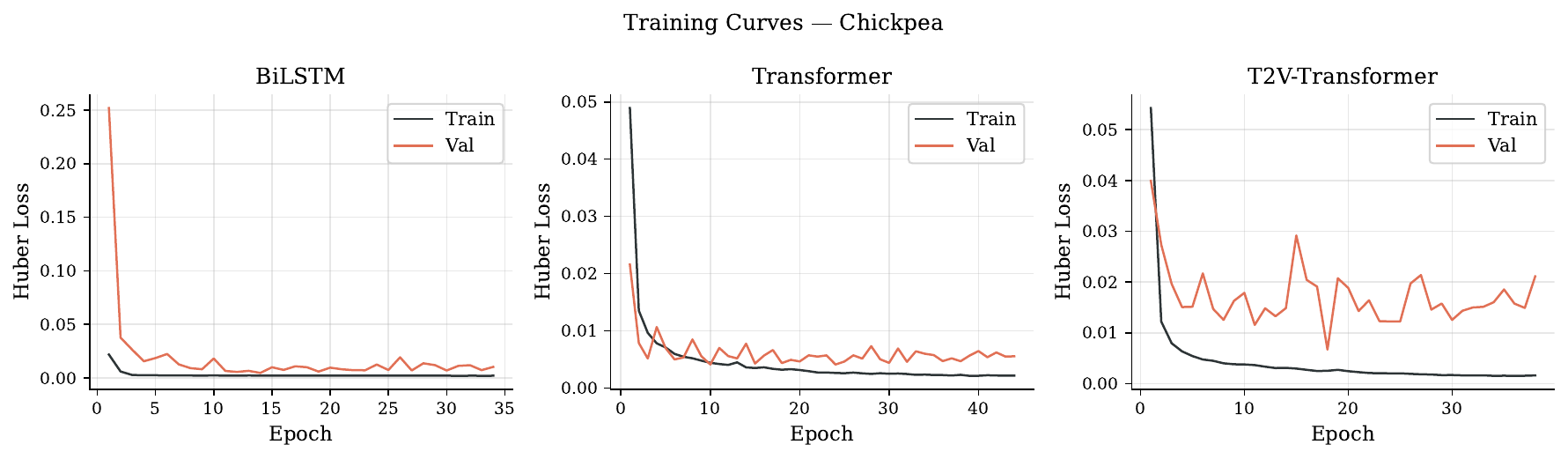}
  \caption{Training curves -- Chickpea. T2V-Transformer shows a persistent train-val gap:
  train loss approaches zero while validation loss plateaus, consistent with learnable
  temporal parameters overfitting the near-random-walk training series.}
  \label{fig:tr_chickpea}
\end{figure}

\begin{figure}[htbp]
  \centering
  \includegraphics[width=\textwidth]{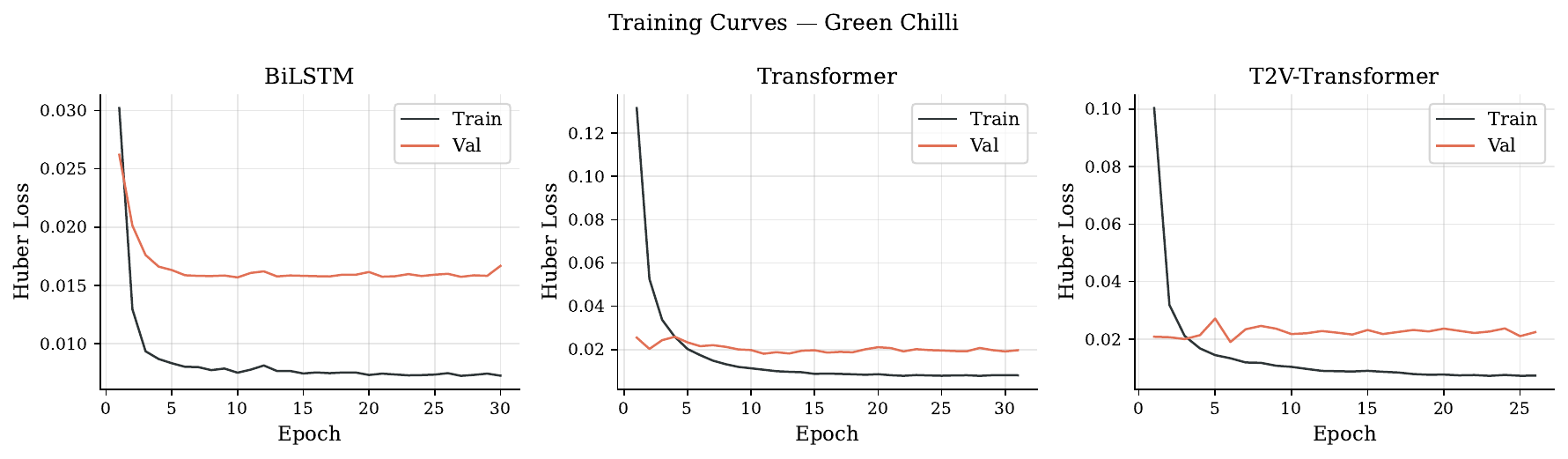}
  \caption{Training curves -- Green Chilli (dropout $=0.3$). Despite increased
  regularisation, persistent train-val gap across all models reflects irreducible noise
  rather than modelling deficiency. Dropout increase prevents further divergence but cannot
  recover signal absent from the training data.}
  \label{fig:tr_chilli}
\end{figure}

\begin{figure}[htbp]
  \centering
  \includegraphics[width=\textwidth]{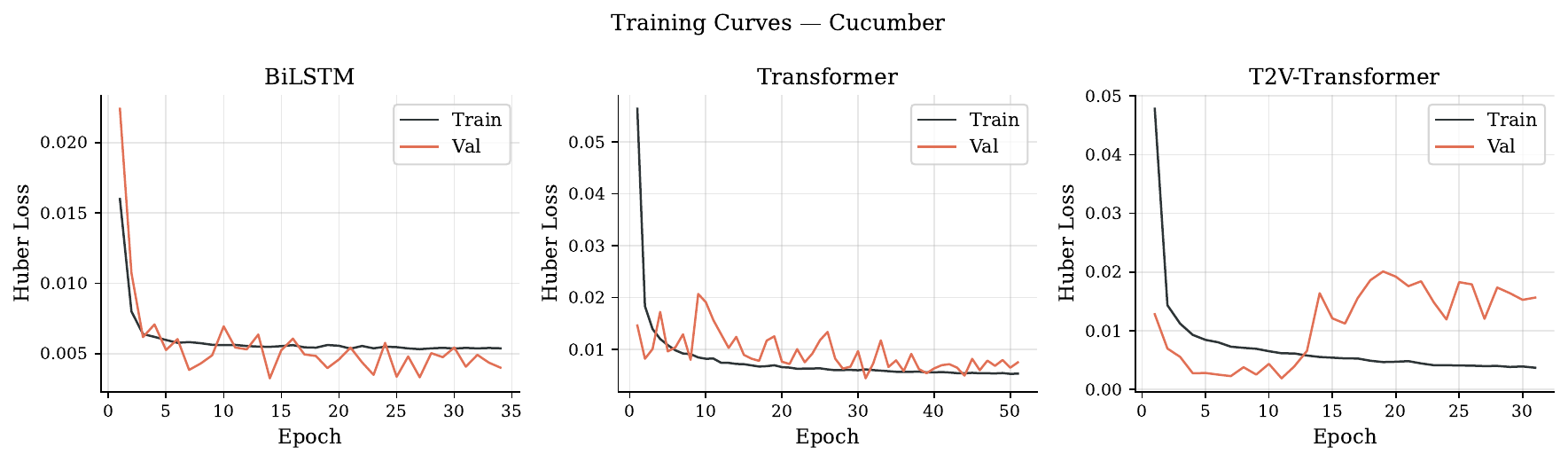}
  \caption{Training curves -- Cucumber. Clean convergence across all models; T2V-Transformer
  validation loss tracks training loss closely, supporting the RMSE improvement observed on
  this commodity.}
  \label{fig:tr_cucumber}
\end{figure}

\begin{figure}[htbp]
  \centering
  \includegraphics[width=\textwidth]{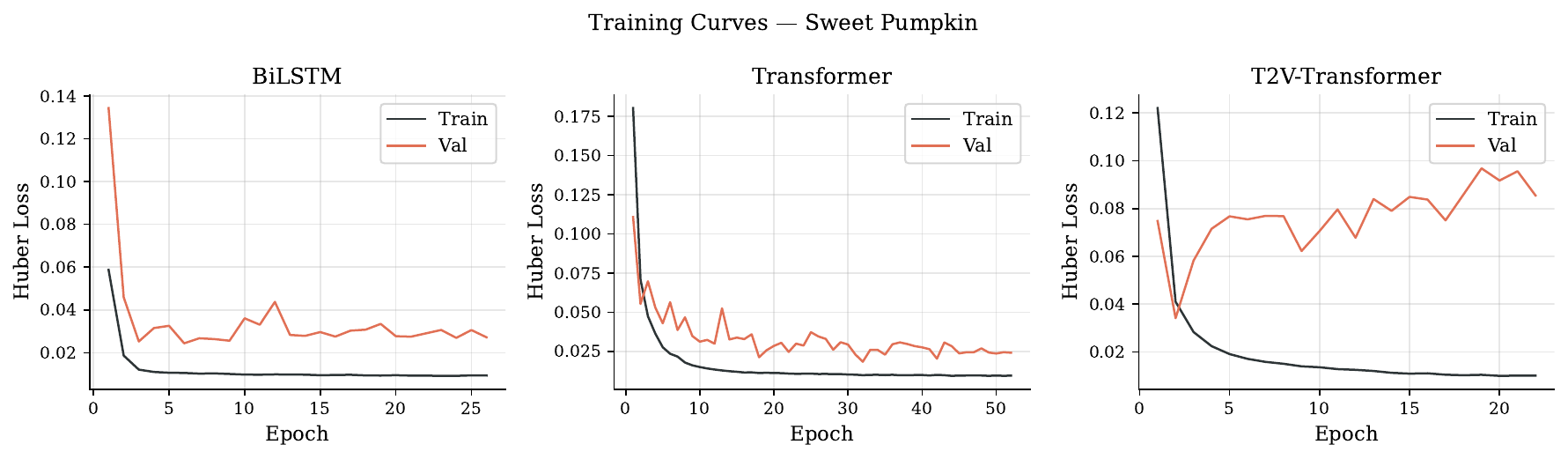}
  \caption{Training curves -- Sweet Pumpkin (dropout $=0.3$). Under default dropout (0.1),
  all deep learning models showed validation loss divergence from early epochs. Increasing
  dropout to 0.3 substantially stabilised training; the Vanilla Transformer achieves the
  best deep learning performance on this commodity (MAE~4.17~BDT/kg).}
  \label{fig:tr_pumpkin}
\end{figure}

\begin{figure}[htbp]
  \centering
  \includegraphics[width=\textwidth]{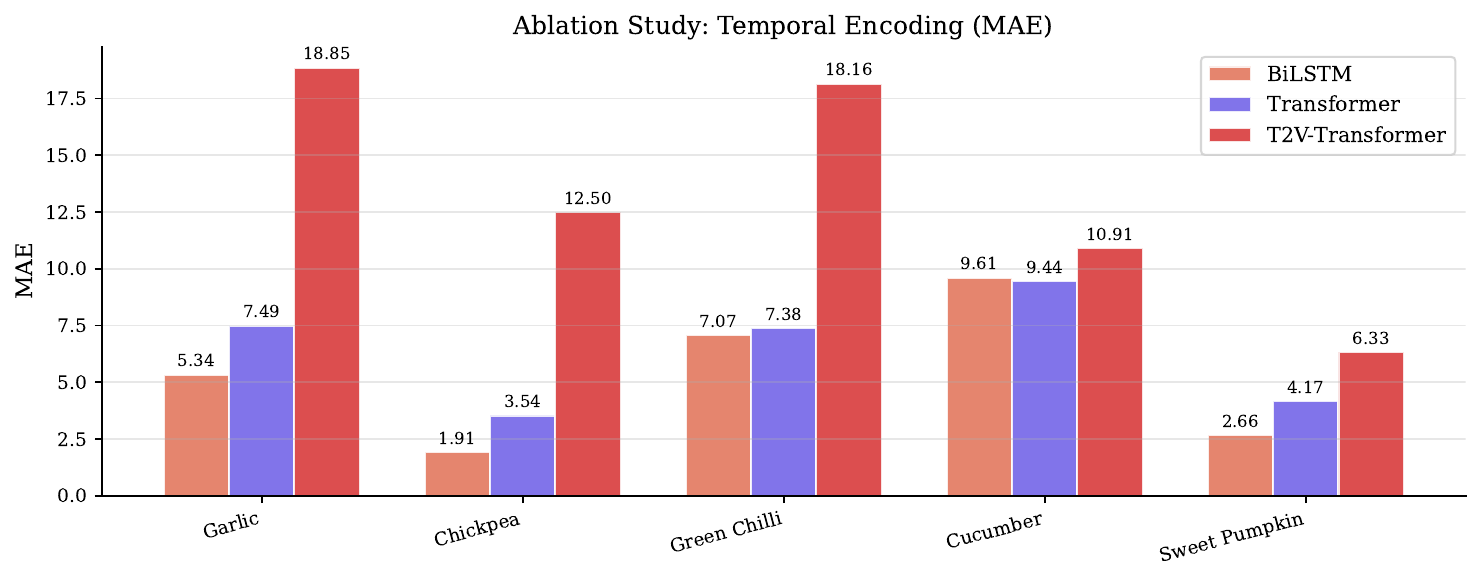}
  \caption{Ablation study -- MAE comparison across BiLSTM, Vanilla Transformer, and
  T2V-Transformer for all five commodities. T2V-Transformer degrades on all five commodities
  by MAE. Garlic ($+151.6\%$) and chickpea ($+253.3\%$) show the most
  extreme degradation alongside green chilli ($+146.1\%$, $p<0.001$).}
  \label{fig:ablation_mae}
\end{figure}

\begin{figure}[htbp]
  \centering
  \includegraphics[width=\textwidth]{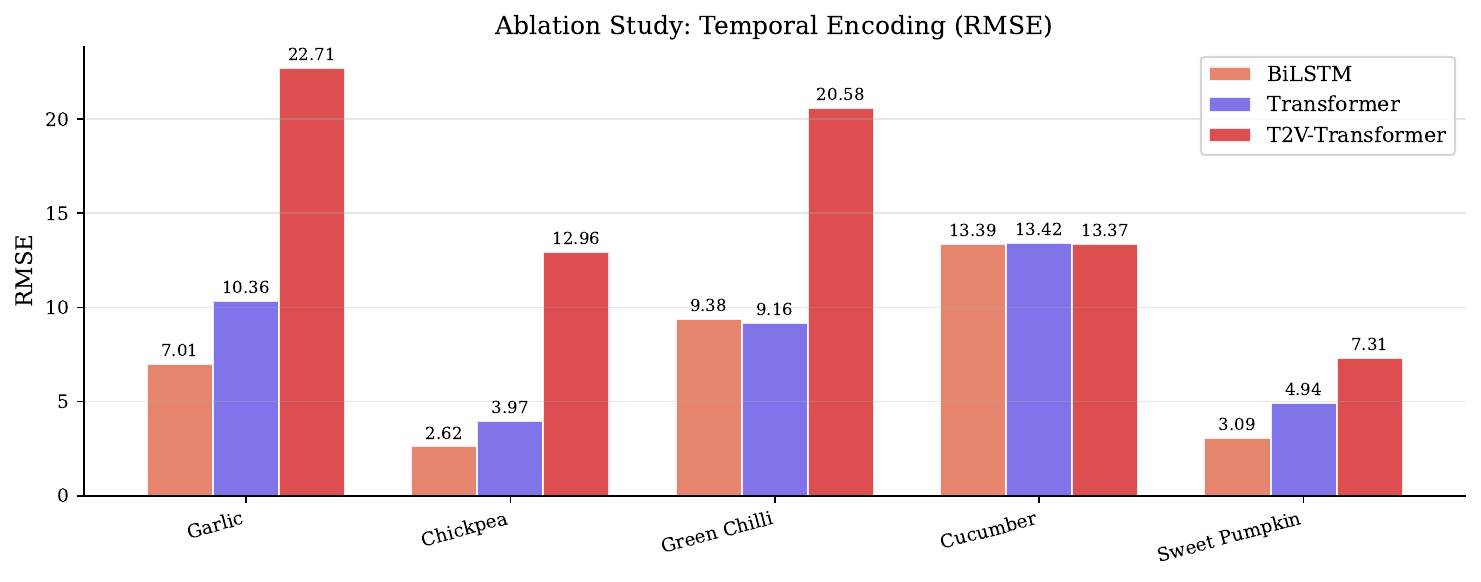}
  \caption{Ablation study -- RMSE comparison across BiLSTM, Vanilla Transformer, and
  T2V-Transformer for all five commodities. The sole RMSE improvement is a modest gain on
  cucumber ($\downarrow$0.4\%), which is not statistically significant ($p=0.962$, DM test,
  Table~\ref{tab:dm}).}
  \label{fig:ablation_rmse}
\end{figure}

\begin{figure}[htbp]
  \centering
  \includegraphics[width=\textwidth]{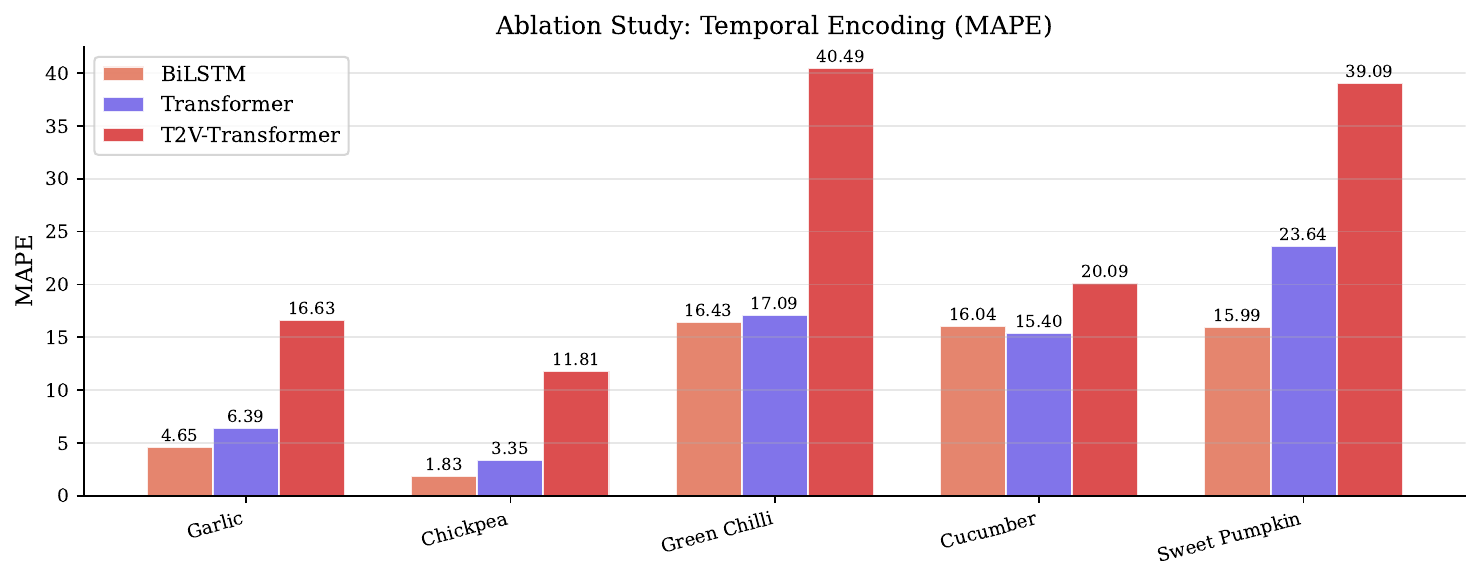}
  \caption{Ablation study -- MAPE comparison across BiLSTM, Vanilla Transformer, and
  T2V-Transformer for all five commodities. Four of five commodities show statistically
  significant Transformer superiority ($p<0.001$), confirming that learnable temporal
  encoding provides no benefit at this training scale.}
  \label{fig:ablation_mape}
\end{figure}

\end{document}